\documentclass[10pt,twocolumn,letterpaper]{article}

\usepackage[pagenumbers]{wacv} 

\usepackage{graphicx}
\usepackage{amsmath}
\usepackage{amssymb}
\usepackage{booktabs}

\usepackage{times}
\usepackage{epsfig}
\usepackage{graphicx}
\usepackage{amsmath}
\usepackage{amssymb}

\usepackage{booktabs}
\usepackage{todonotes}
\usepackage{tikz}
\usepackage{pgfplots}
\usetikzlibrary{shapes.arrows, fadings}
\usetikzlibrary{arrows}
\usetikzlibrary{3d,calc}
\usepackage{comment}
\usepackage{pifont}
\newcommand{\cmark}{\ding{51}}
\newcommand{\xmark}{\ding{55}}

\usepackage{caption} 
\definecolor{myred}{rgb}{0.7,0.25,0.2}
\definecolor{mygreen}{RGB}{0, 153, 0}
\definecolor{myblue}{RGB}{0, 153, 255}
\definecolor{myorange}{RGB}{255, 153, 51}
\definecolor{mygray}{rgb}{0.4,0.4,0.4}
\definecolor{Green}{RGB}{210, 253, 210}
\definecolor{LightGreen}{RGB}{236, 255, 239}

\pgfplotstableread[col sep = comma]{./csv/Number_years.csv}\dataYearsA
\pgfplotstableread[col sep = comma]{./csv/Observation_years1.csv}\dataYearsB
\pgfplotstableread[col sep = comma]{./csv/Observation_years2.csv}\dataYearsC
\pgfplotstableread[col sep = comma]{./csv/Turtle_counts.csv}\dataTurtleCounts
\pgfplotstableread[col sep = comma]{./csv/Rotation_1.csv}\dataKeypointsA
\pgfplotstableread[col sep = comma]{./csv/Matches_5_0.csv}\dataKeypointsB
\pgfplotstableread[col sep = comma]{./csv/Matches_5_1.csv}\dataKeypointsC
\pgfplotstableread[col sep = comma]{./csv/Transitions_mod.csv}\dataTransitions
\pgfplotstableread[col sep = comma]{./csv/Matches_time.csv}\dataMatchesTime
\pgfplotstableread[col sep = comma]{./csv/NewIdentities.csv}\newIdentities
\pgfplotstableread[col sep = comma]{./csv/random_head_knn.csv}\knnHeadRandom
\pgfplotstableread[col sep = comma]{./csv/timeaware_head_knn.csv}\knnHeadTimeaware
\pgfplotstableread[col sep = comma]{./csv/random_turtle_knn.csv}\knnTurtleRandom
\pgfplotstableread[col sep = comma]{./csv/timeaware_turtle_knn.csv}\knnTurtleTimeaware

\usetikzlibrary{math}
\usetikzlibrary{fit}

\tikzset{%
nodeA/.style={circle,draw,black,fill=blue,inner sep=0pt,minimum size=6pt},
nodeB/.style={rectangle,draw,black,fill=orange,inner sep=0pt,minimum size=6pt},
nodeC/.style={circle,draw,white,fill=white,inner sep=0pt,minimum size=0pt},
container/.style={draw, dashed, rectangle, inner sep=0.3cm, rounded corners}
}

\usepackage[pagebackref=true,breaklinks=true,letterpaper=true,colorlinks,bookmarks=false]{hyperref}

\usepackage[capitalize]{cleveref}
\crefname{section}{Sec.}{Secs.}
\Crefname{section}{Section}{Sections}
\Crefname{table}{Table}{Tables}
\crefname{table}{Tab.}{Tabs.}

 \usepackage{multirow}
 \usepackage{booktabs,adjustbox}

\def\wacvPaperID{1248} 
\def\confName{WACV}
\def\confYear{2024}

\begin{document}

\title{SeaTurtleID2022: A long-span dataset for reliable sea turtle re-identification}

\author{Lukáš Adam$^\dagger$, Vojtěch Čermák$^\dagger$\\
Czech Technical University\\
{\tt\small lukas.adam.cr@gmail.com} \\
{\tt\small cermavo3@fel.cvut.cz}
\and
Kostas Papafitsoros$^\dagger$\\
Queen Mary University of London\\
{\tt\small k.papafitsoros@qmul.ac.uk}
\and
Lukas Picek$^\dagger$\\
UWB and INRIA\\
{\tt\small picekl@kky.zcu.cz}\\ 
{\tt\small lpicek@inria.fr}
\and
{\tt\small $\dagger$ - Equal contribution in alphabetical order}
}
\maketitle
\begin{abstract}
This paper introduces the first public large-scale, long-span dataset with sea turtle photographs captured in the wild -- \href{https://www.kaggle.com/datasets/wildlifedatasets/seaturtleid2022}{SeaTurtleID2022}. The dataset contains 8729 photographs of 438 unique individuals collected within 13 years, making it the longest-spanned dataset for animal re-identification. All photographs include various annotations, e.g., identity, encounter timestamp, and body parts segmentation masks. 
Instead of standard "random" splits, the dataset allows for two realistic and ecologically motivated splits: (i) a \textit{time-aware closed-set} with training, validation, and test data from different days/years, and (ii) a \textit{time-aware open-set} with new unknown individuals in test and validation sets.
We show that time-aware splits are essential for benchmarking re-identification methods, as random splits lead to performance overestimation.
Furthermore, a baseline instance segmentation and re-identification performance over various body parts is provided. 
Finally, an end-to-end system for sea turtle re-identification is proposed and evaluated. The proposed system based on Hybrid Task Cascade for head instance segmentation and ArcFace-trained feature-extractor achieved an accuracy of 86.8\%.
\end{abstract}

\begin{figure}[t]
\centering
\begin{minipage}[t]{0.19\textwidth}
\centering
\includegraphics[width=0.95\textwidth]{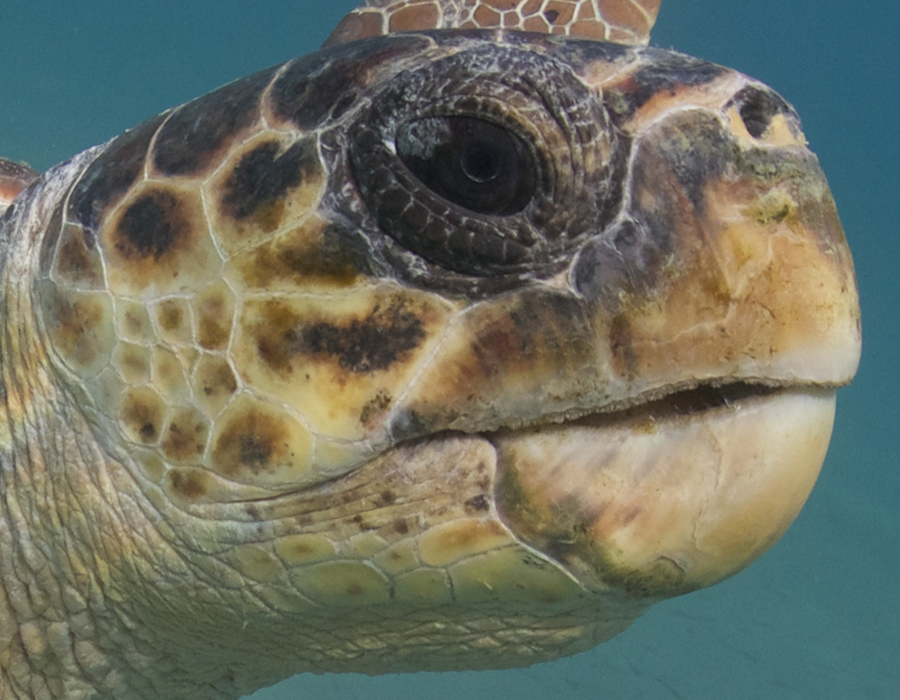}\\[0.2em]
\includegraphics[width=0.95\textwidth]{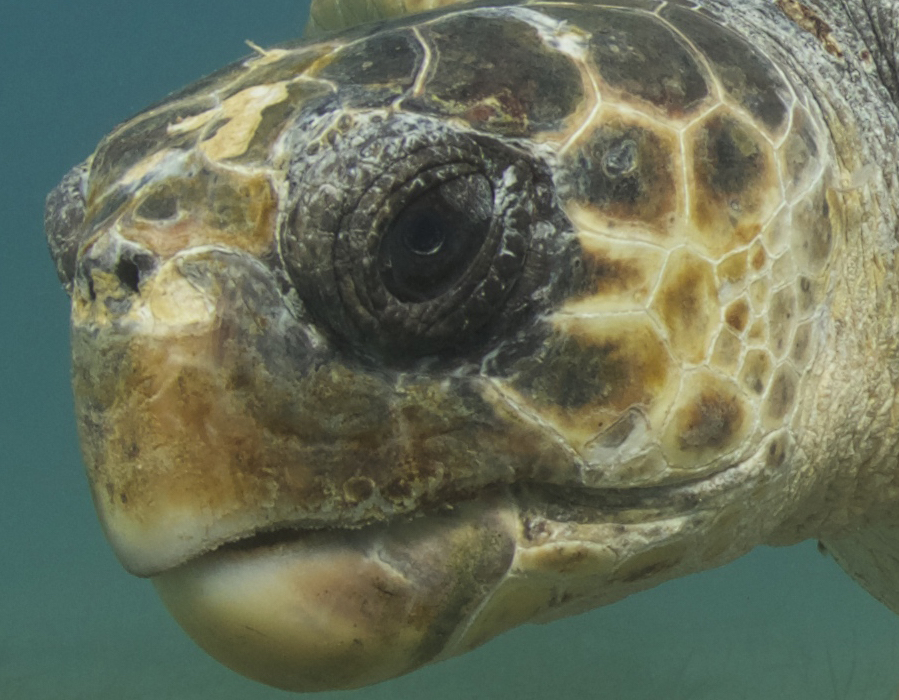}\\[0.2em]
\small{\textbf{2014}}
\end{minipage}
\begin{minipage}[c]{0.05\textwidth}
\begin{center}
\vspace{0.5cm}
$\Large{\boldsymbol{\rightarrow}}$ \\
\vspace{2.1cm} 
$\Large{\boldsymbol{\rightarrow}}$
\end{center}
\end{minipage}
\begin{minipage}[t]{0.19\textwidth}
\centering
\includegraphics[width=0.95\textwidth]{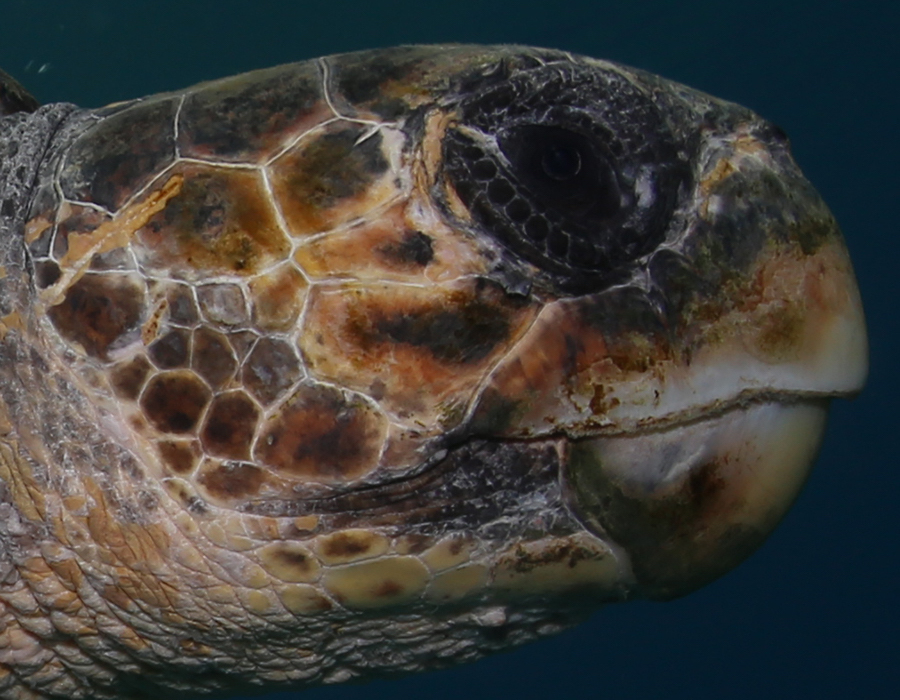}\\[0.2em]
\includegraphics[width=0.95\textwidth]{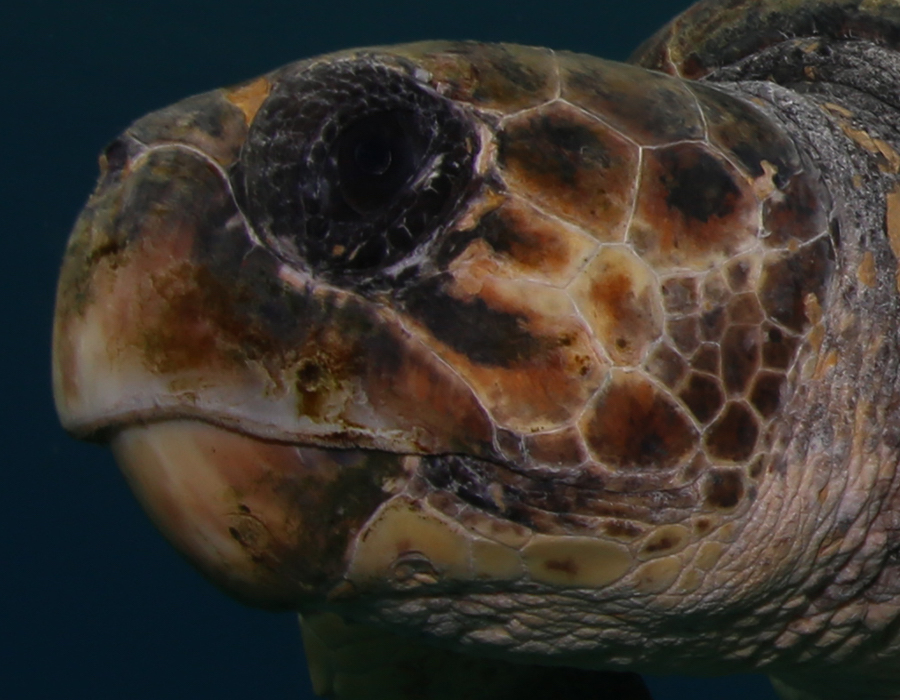}\\[0.2em]
\small{\textbf{2021}}
\end{minipage}

\caption{The long-span difference in visual appearance of one individual sea turtle. The shapes of the facial scales remain the same, but other features, e.g., coloration, pigmentation, shape, and scratches, change over time.}
\label{fig:change_patterns}
\vspace{-0.25cm}
\end{figure}

\vspace{-0.25cm}
\section{Introduction}
Image-based individual animal re-identification, i.e., the process of recognizing individual animals based on their unique stable-over-time external characteristics, is essential for studying different aspects of wildlife, like population monitoring, movements, behavioral studies, and wildlife management\,\cite{papafitsoros_2021,Schofield_2022,vidal2021perspectives}. The increasing sizes of the associated photo databases stemming from the multi-year span of such projects\,\cite{Schofield_2020,swanson2015snapshot} have highlighted the need for automated methods to reduce labor-intensive human supervision in individual animal identification.

As a result, a plethora of automatic re-identification methods have been developed during the last years\,\cite{blount2022flukebook,crall2013hotspotter,korschens2019elpephants,weideman2020extracting}. Evaluation of these methods is performed on benchmark databases, covering several animal groups like mammals\,\cite{belugaid,li2019atrw,nepovinnykh2022sealid,trotter2020ndd20,botswana2022,zuffi2019three}, reptiles\,\cite{zinditurtles, crall2013hotspotter}, and smaller organisms\,\cite{bruslund2020re,ferreira2020deep,schneider2018can}. Typically, such databases are split into a \textit{reference set} -- a set of images with individuals whose identity (label) is known -- and a \textit{query set} -- the set of images with individuals whose identity needs to be matched to the reference set. 
In deep learning, these sets are usually called training and test sets.

The quality of datasets influences the objectivity of the method evaluation. Therefore, the dataset and its splitting should mimic a realistic scenario, i.e., the images in the \textit{query} and \textit{reference} sets should not originate from the same \textit{encounters} (burst mode in camera traps, consecutive video frames, multiple photographs taken during an encounter) and/or capture unknown identities.
Other \textit{factors}, e.g., different locations, image capture conditions, and images that reflect changes in animal appearances over time, are also vital. For reference see Figure\,\ref{fig:change_patterns}.

Typically, images produced during one \textit{encounter} share the same factors as the encounter lasts for a short period. The most efficient way to indicate different encounters and factors in a dataset is by including the capture date and time in metadata, i.e., \textit{timestamps}.
Without knowing the time of the observation, datasets are often split into reference and query sets exclusively randomly. Therefore, images in training and test sets often originate from the same encounter/observation, representing unwanted training-to-test data leakage.
This might result in overfitting to factors of a particular encounter instead of learning an inner representation of each individual.
Thus, a random split implicitly assumes that one will encounter the same factors in the future, which is highly unrealistic.
On the other hand, timestamps allow for time-aware splits, where images from a time period are all in either the reference or the query set. This leads to a more realistic case in which new factors are encountered in the future.

Based on our extensive research, just five publicly available datasets contain timestamps (see Table\,\ref{table:datasets}). From those, Cows2021\,\cite{gao2021towards} and GiraffeZebraID\,\cite{parham2017animal} span only one month, and WhaleSharkID\,\cite{holmberg2009estimating} includes timestamps for only 9\% of photographs. This leaves only two wildlife datasets with timestamps; with span of at most two years.
We introduce a novel dataset with photographs of loggerhead sea turtles (\textit{Caretta caretta}) -- the SeaTurtleID2022. The dataset was collected over 13 years and consists of 8729 high-resolution photographs of 438 unique individuals. Each photograph includes various annotations, e.g., identities, encounter timestamps, and body parts segmentation masks. To the best of our knowledge, the SeaTurtleID2022 is the longest-spanned public wild animal image dataset and the only public dataset of sea turtles with photographs captured in the wild. In contrast to existing datasets, the SeaTurtleID2022 allows for two realistic and ecologically motivated splits instead of a  \textit{"random"} split:
\begin{itemize}
 \vspace{-0.1cm}
\itemsep0pt
    \item \textit{time-aware closed-set}: with reference images belonging to different encounters than query ones, and
    \item \textit{time-aware open-set}: with new \textit{unknown} individuals (i.e., newly introduced to population) in test and validation sets (common in ecology).

\end{itemize}

\begin{table}[!hb]
\vspace{-0.15cm}
\setlength{\tabcolsep}{0.45em} 
\small
\centering
\begin{tabular}{@{}l@{}rrrrr@{}}
\toprule
        \textbf{Dataset} &  \textbf{images} & \textbf{t-stamp} & \textbf{ind.} &  \textbf{enc.} &       \textbf{span} \\
\midrule
       Cows2021\,\cite{gao2021towards} & 8670 & 100\% & 181 & 3036 & 31 \\
 GiraffeZebraID\,\cite{parham2017animal} &             6925 &          100\% &             2051 &           2494 &    12  \\
   MacaqueFaces\,\cite{witham2018automated} &             6280 &          100\% &               34 &            494 &   525  \\
    BelugaID\,\cite{belugaid} &             5902 &          100\% &              788 &           1241 &   785 \\
   WhaleSharkID\,\cite{holmberg2009estimating} &             7693 &            9\% &               98 &            424 &  1971 \\
\midrule
   SeaTurtleID2022 (ours) &             8729 &          100\% &              438 &           1221 &  4390  \\
\bottomrule
\end{tabular}
\caption{Dataset statistics for all publicly available animal re-identification datasets with timestamps; number of photographs, percentage of photographs with timestamps, number of individuals and encounters, and dataset span in days.
}
\label{table:datasets}
\end{table}

Even though the SeaTurtleID2022 dataset is intended primarily as an animal re-identification benchmark, it can be used for the evaluation and testing of several fundamental problems, including:
(i) object detection,
(ii) instance segmentation,
(iii) fully- and weakly supervised semantic segmentation,
(iv) 3D reconstruction, and
(v) concept drift analysis.

We stress that SeaTurtleID2022 lacks common drawbacks of other (human) re-identification datasets. In particular, face-id datasets typically contain low-resolution photographs, are restricted to limited poses, have limited time spans, and are either artificially generated\,\cite{Bae_2023_WACV}, or collected by crawling the internet\,\cite{LFWTechUpdate}, raising privacy concerns.

Apart from the dataset, we provide a baseline performance evaluation for body-part segmentation and re-identification.
Based on that, a baseline methodology for wildlife re-identification is proposed and evaluated over the SeaTurtleID2022 and several other well-known datasets using hand-crafted features and metric learning approaches.
The best ArcFace-trained feature extractor achieved an accuracy of 69.2\% on the SeaTurtleID2022 dataset while using cropped heads. In case no body part detection is done, the use of full images and the same approach resulted in an accuracy of 17.1\%, showing that turtle identification is still a challenging task without body parts detection.

Furthermore, we showcase that time-unaware splits can often lead to performance overestimation if compared to time-aware splits. Hence, we recommend evaluating re-identification-focused algorithms over datasets with timestamps and unbiased (e.g., time-aware) splits. Additionally, imaging data collectors and database curators should ensure that time information is included in the metadata.

\begin{figure*}[!t]
\centering
\begin{minipage}[t]{0.32\textwidth}
\centering
\includegraphics[width=0.95\textwidth]{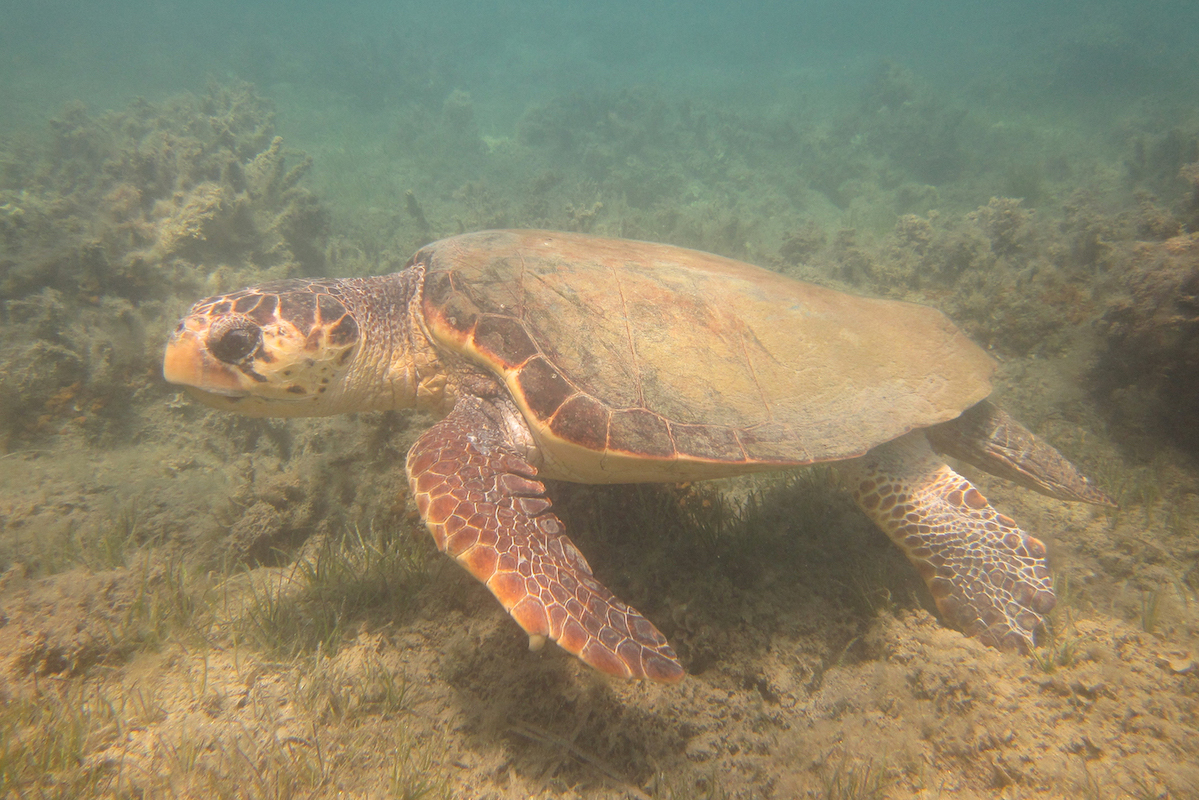}\\[0.2em]
\small{\textbf{2011}: compact camera, no flash}
\end{minipage}
\begin{minipage}[t]{0.32\textwidth}
\centering
\includegraphics[width=0.95\textwidth]{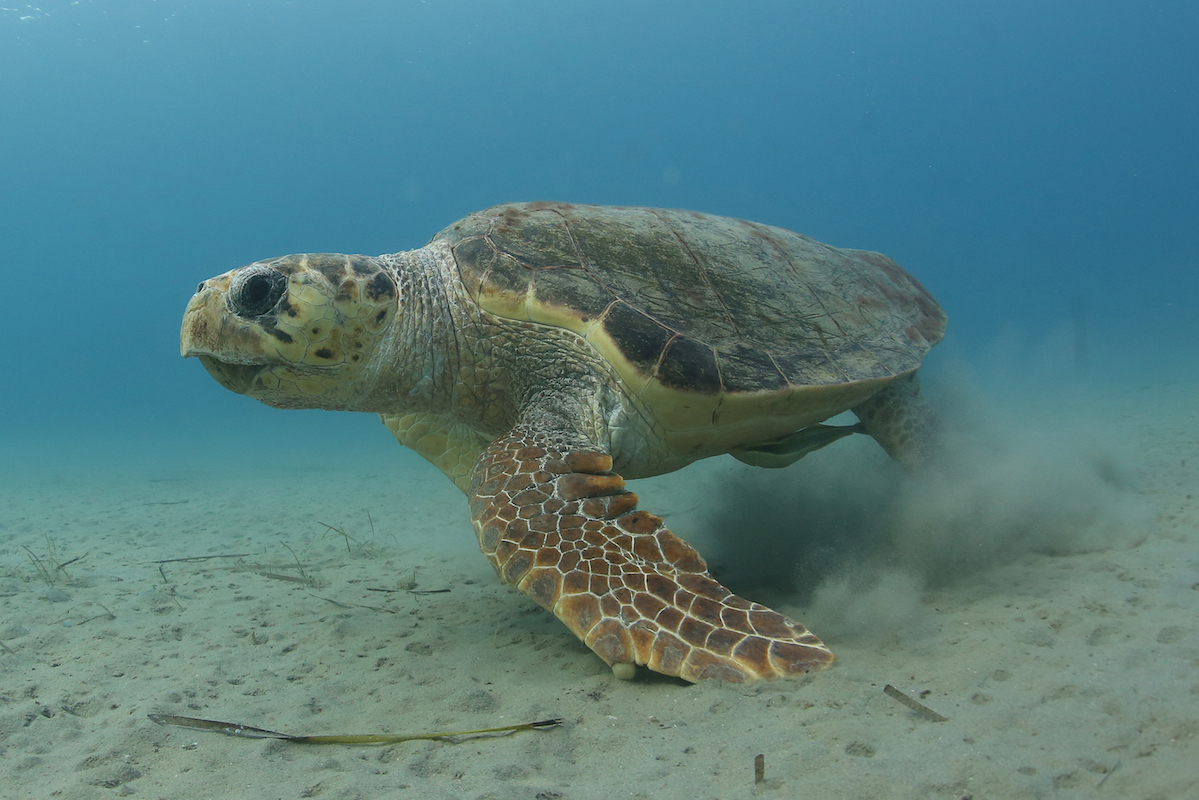}\\[0.2em]
\small{\textbf{2014}: DSLR camera, no flash}
\end{minipage}
\begin{minipage}[t]{0.32\textwidth}
\centering
\includegraphics[width=0.95\textwidth]{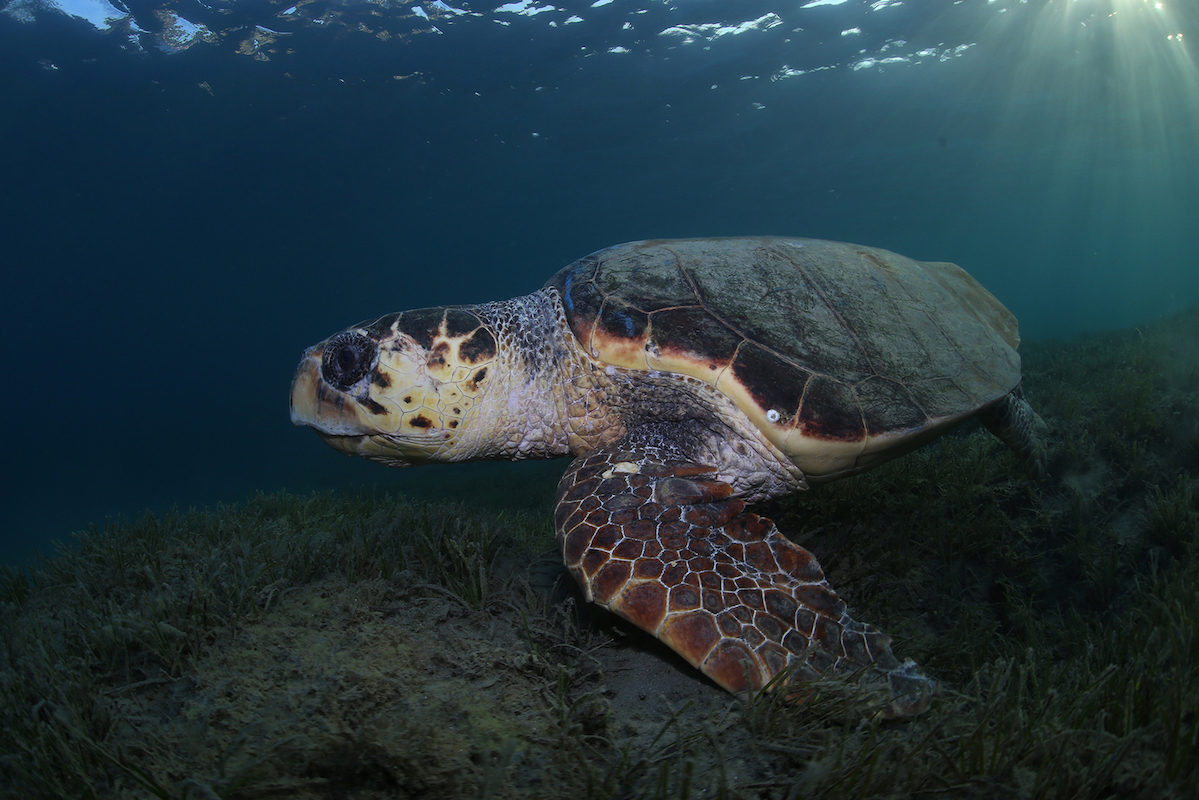}\\[0.2em]
 \small{\textbf{2019}: DSLR camera, with flash}
\end{minipage}
\caption{Selected individual turtle (t023) from the SeaTurtleID2022 database, photographed with three different camera set-ups. Photographs taken with the DSLR camera are of higher quality, and the additional use of flash recovers the natural colouration of the animal. All the photographs were cropped for illustration purposes. }
\label{fig:t023}
\end{figure*}

The main contributions of this paper are as follows:
\begin{itemize}
\itemsep0pt
    \item We introduce a novel dataset -- \href{https://www.kaggle.com/datasets/wildlifedatasets/seaturtleid2022}{SeaTurtleID2022} -- for animal re-identification with unique characteristics and a wide variety of annotations, e.g., identities, encounter timestamps,  segmentation masks, bounding boxes, and orientations for all body parts.
    \item We provide (i) baseline re-identification performance evaluation using hand-crafted features and metric learning approaches over SeaTurtleID2022 and other established datasets and (ii) baseline performance for body-part segmentation using well-known instance segmentation methods.
    \item We provide empirical evidence that a time-unaware splitting of the dataset leads to a significant overestimation bias.
    \item Based on all the above, we have developed and evaluated an end-to-end system for reliable sea turtle identification in the wild that can potentially be transferred to other species as well.
\end{itemize}

\newpage
\section{The SeaTurtleID2022 dataset}

This section describes the data collection process, annotation procedures, and key features of the dataset.

\subsection{Data collection}

\noindent\textbf{Location and species}: All photographs were taken in Laganas Bay, Zakynthos Island, Greece (37$^{\circ}$43$'$N, 20$^{\circ}$52$'$E), from 2010 until 2022; May--October. Laganas Bay is a main breeding site for the Mediterranean loggerhead sea turtles\,\cite{Marga_Pana}. Female turtles (around 300 annually) are mainly migratory and visit the island to breed every 2--3 years\,\cite{Schofield_2020}. On the other hand, certain individuals reside on the island, and they can be observed in consecutive years\,\cite{papafitsoros_2021,Schofield_2022}. Loggerheads are long-lived species, and they can have reproductive longevity of more than three decades\,\cite{longevity}, which can lead to long-span image recordings for specific individuals. Sea turtles are particularly amenable to photo-identification due to their scale patterns\,\cite{Schofield_ID}. In particular, the polygonal scales in the lateral (side) and dorsal (top) sides of their heads are unique to every individual and remain stable throughout their lives\,\cite{Carpentier_2016}, see Figure\,\,\ref{fig:change_patterns} and additional examples in the supplementary material. Notably, the left and right side patterns differ for a given turtle. \vspace{-0.2cm} \\

\noindent\textbf{Photographic procedure}: All photographs were captured underwater during snorkeling surveys from a distance ranging from 7 meters to a few centimeters using three cameras: (i) Canon IXUS 105 digital compact camera with a Canon underwater housing in 2010--2013, (ii) Canon 6D full-frame DSLR camera combined with a Sigma 15mm fisheye lenses and an Ikelite underwater housing in 2014--2017, and (iii) the same camera with an additional INON Z330 external flash in 2018--2022. The resolution ranges from 4000$\times$3000 (Canon IXUS) to 5472$\times$3648 pixels (Canon 6D) with an average of 5269$\times$3564. The water depth ranged from 1 to 8 meters, with the vast majority of photographs taken less than 5 meters deep.

Photographs taken in 2014--2022 are generally of better quality due to the use of a more advanced camera and a shorter camera-subject distance. On the other hand, due to the use of fisheye lenses, barrel shape distortion can be noticeable, especially for close-up photographs. Finally, more natural colors were acquired using the external flash. In Figure\,\ref{fig:t023}, we display three images of the same individual -- obtained by the three different camera set-ups -- to highlight the resulting visual differences.

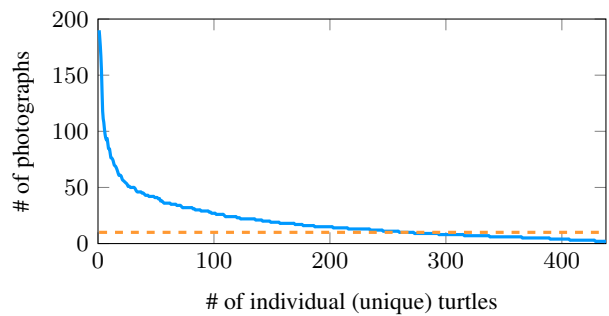
\begin{figure}[!b]
 \centering
\begin{tikzpicture}[font=\small]
\begin{axis}[
      width=\linewidth,
      xmin=0,
      xmax=438,
      ymin=0,
      ymax=200,
      title={},
      xlabel={\# of individual (unique) turtles},
      ylabel={\# of photographs},
      ylabel near ticks,
      height=130,
]
      \addplot [mark=none, very thick, smooth, myblue] table[x index=0, y index=1] {\dataTurtleCounts}; 
      \addplot [dashed, very thick, myorange] (1,10) -- (450,10);
\end{axis}
\end{tikzpicture}
\caption{Number of photographs for each of the 438 turtles. The orange line corresponds to 10 photographs.}
 \label{fig:individuals}
\end{figure}

\begin{figure*}[!ht]
\begin{minipage}[t]{0.36\linewidth}
 \centering
\begin{tikzpicture}[font=\footnotesize]
\begin{axis}[
      width=\linewidth,
      xmin=2009.5,
      xmax=2022.5,
      ymin=0,
      ymax=270,
      title={},
      xlabel={\footnotesize{year}},
      ylabel={\footnotesize{\# of encounters}},
      xtick={2010,2012,2014,2016,2018,2020,2022},
      xticklabels={2010,2012,2014,2016,2018,2020,2022},
      ytick={0,50,100,150,200,250},
      yticklabels={0,50,100,150,200,250},
      xtick pos=left,
      ylabel near ticks,      
      height=100,
      ybar,
      bar width=2.5mm,
]
      \addplot table[x index=0, y index=2] {\dataYearsA}; 
\end{axis}
\end{tikzpicture}
\end{minipage}
\hfill
\begin{minipage}[t]{0.27\linewidth}
\begin{tikzpicture}[font=\footnotesize]

\begin{axis}[
      width=\linewidth,
      xmin=0,
      xmax=8.5,
      ymin=0,
      ymax=270,
      title={},
      xlabel={\footnotesize{\# of observation years}},
      ylabel={\footnotesize{\# of individuals}},
      xtick={1,2,3,4,5,6,7,8,9,10},
      ytick={0,50,100,150,200,250},
      xtick pos=left,
      ylabel near ticks,
      height=100,
      ybar,
      bar width=2.5mm,
]
      \addplot table[x index=0, y index=1] {\dataYearsB}; 
\end{axis}
\end{tikzpicture}
\end{minipage}
\hfill
\begin{minipage}[t]{0.36\linewidth}
\begin{tikzpicture}[font=\footnotesize]
\begin{axis}[
      width=\linewidth,
      xmin=2009.5,
      xmax=2022.5,
      ymin=0,
      ymax=110,
      title={},
      xlabel={\footnotesize{year}},
      ylabel={\footnotesize{\# of new identities}},
      xtick={2010,2012,2014,2016,2018,2020, 2022},
      xticklabels={2010,2012,2014,2016,2018,2020, 2022},
      ytick={0,20,40,60,80,100},
      ytick={0,20,40,60,80,100},
      xtick pos=left,
      ylabel near ticks,      
      height=100,
      ybar,
      bar width=2.5mm,
]
      \addplot table[x index=0, y index=1] {\newIdentities}; 
\end{axis}
\end{tikzpicture}
\end{minipage}
\caption{Time-related statistics within the SeaTurtleID2022 dataset: number of encounters per year (left),  distribution of all individuals to the total number of observation years, i.e., recurrence of individuals (middle), and number of newly observed identities in each year (right).}
\label{fig:time}
\end{figure*}
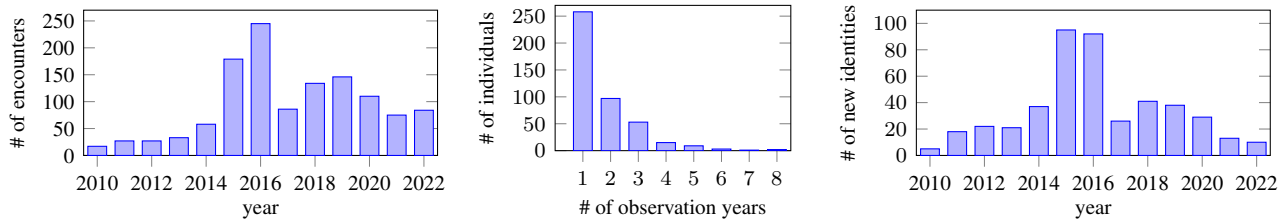

\subsection{Dataset highlights}

\noindent\textbf{Large-scale in the wild dataset}: With 8729 photographs and 438 individuals, the dataset represents the most extensive publicly available dataset for sea turtle identification in the wild. The images are in original resolution and with various backgrounds. Approximately 90\% of photographs have a size of 5472$\times$3648 pixels, the average photograph size is 5269$\times$3564 pixels, while the head occupies on average 635$\times$554 pixels. Figure\,\ref{fig:individuals} shows the number of photographs for each individual. The majority of individuals ($\frac{272}{438}$) have at least ten photographs (depicted by the dashed line). Similarly, most individuals ($\frac{270}{438}$) were encountered at least twice. We note that this number is expected to increase in the following years since this dataset is updated annually.

\noindent\textbf{Long time span \& timestamps}: The dataset contains photographs continuously captured over 13 years from 2010 to 2022. In contrast to most existing animal datasets that are usually collected in controlled environments and/or over a short time span, the SeaTurtleID2022 dataset includes a timestamp (in dd:mm:yyyy format) for each photograph.
Figure\,\,\ref{fig:time} (left) shows the number of encounters for each year, with a significantly larger number from 2015 onwards. We note that this is driven by an increasing data collection effort rather than reflecting actual annual recurrence. In Figure\,\,\ref{fig:time} (right), we show the number of newly observed individuals.
Furthermore, Figure\,\,\ref{fig:time} (middle) shows the distribution of the 438 individuals with respect to the total number of observation years. A span of one year means that a turtle was photographed only in one year. Many turtles ($\frac{180}{438}$) were photographed in at least two different years, and 9 individual turtles spanned more than 9 years. \vspace{-0.25cm} \\

\noindent\textbf{Segmentation masks and bounding boxes}: Almost all photographs in the dataset have a visible head and/or flippers. Therefore we provide body parts annotations photographs as segmentation masks and bounding boxes. Apart from masks, we include orientation (left, right, top, top-right, top-left, front or bottom) for each head mask, and orientation (top or bottom) and location (front left/right or rear left/right) for flipper masks. 
Such annotations allow further development and evaluation of turtle identification methods or novel methods for object detection and semantic segmentation. 
All segmentation mask annotations were done semi-automatically using the Segment Anything\,\cite{kirillov2023segment} model integrated within the \href{https://github.com/opencv/cvat}{CVAT}. 
\vspace{-0.25cm} \\

\noindent\textbf{Multiple poses}: The dataset includes multiple images from different angles and, therefore, provides a ground for the challenging task of 3D animal reconstruction. \vspace{-0.25cm} \\

\noindent\textbf{Comparison with ZindiTurtleRecall}\,\cite{zinditurtles}: For a better perspective, we compare the SeaTurtleID2022 with the ZindiTurtleRecall dataset, which is the only other publicly available sea turtle dataset. We stress that the latter dataset contains photographs in a controlled environment (a rehabilitation center) with no timestamps. We summarise all comparable aspects of both datasets in Table\,\ref{table:dataset}.

\begin{figure}[!h]
\vspace{0.15cm}
 \centering
 \includegraphics[width=0.48\linewidth, height=2.8cm]{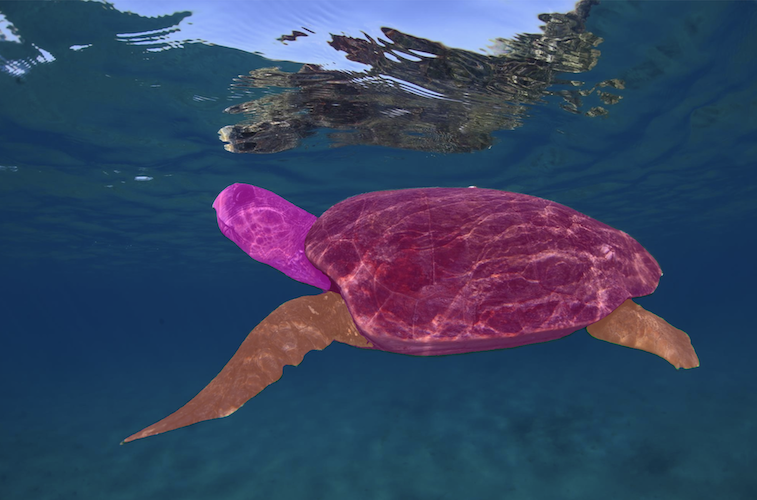}
 \includegraphics[width=0.48\linewidth, height=2.8cm]{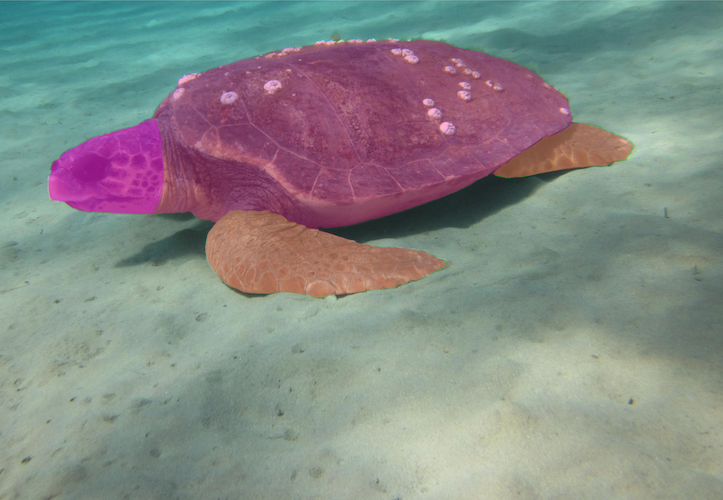} \\
 \vspace{0.05cm}
  \includegraphics[width=0.48\linewidth, height=2.8cm]{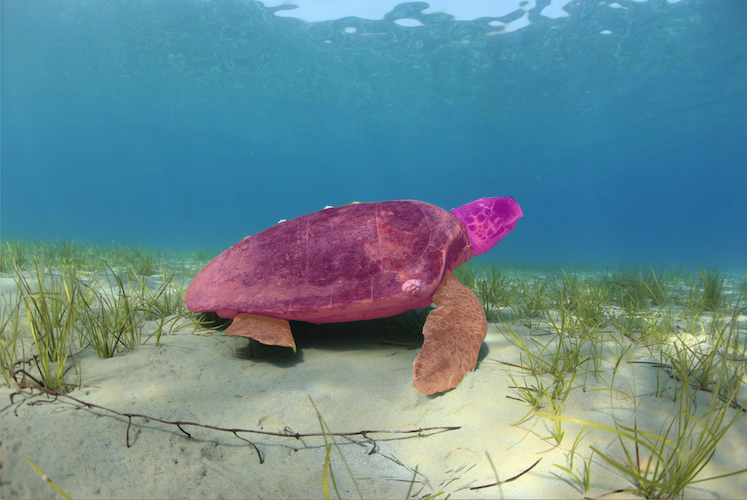}
 \includegraphics[width=0.48\linewidth, height=2.8cm]{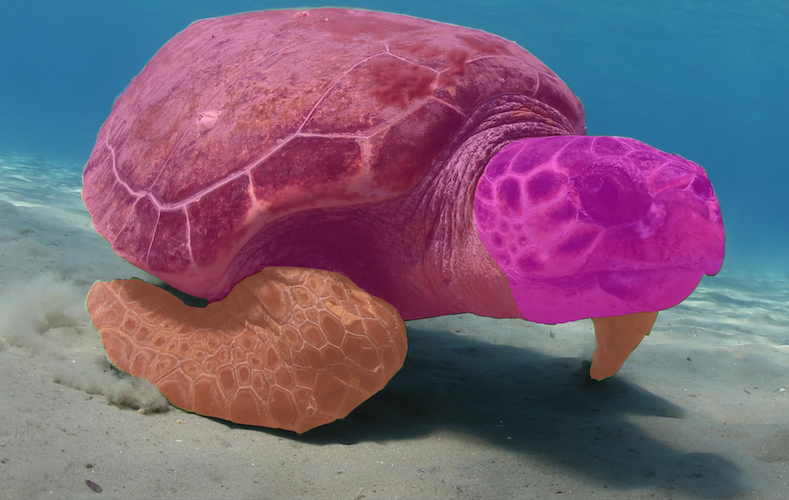} \\
 \vspace{0.05cm}
  \includegraphics[width=0.48\linewidth, height=2.8cm]{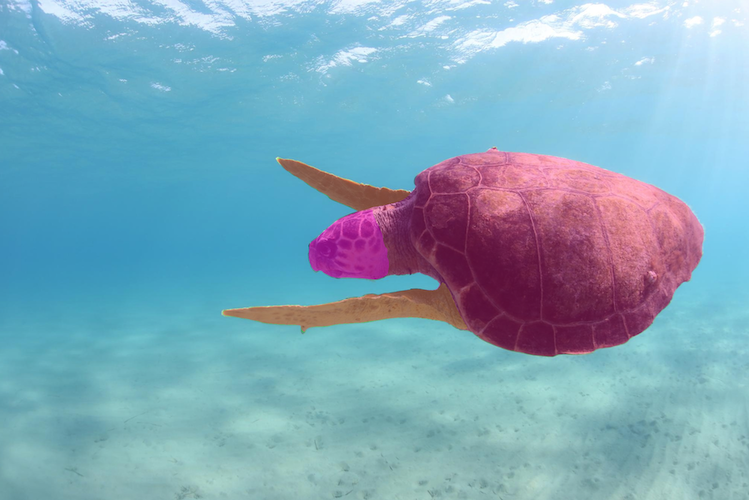}
 \includegraphics[width=0.48\linewidth, height=2.8cm]{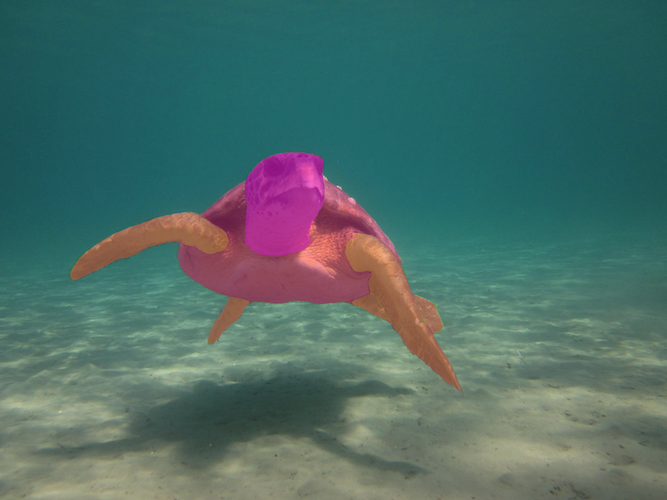}
 \caption{Examples of body parts (head, carapace, flippers) segmentation masks.}
 \label{fig:annotation}
\end{figure}

\begin{table}[!h]
\setlength{\tabcolsep}{0.3em} 

\small
\centering
\begin{tabular}{@{}lrr@{}}
 \toprule
 & \textbf{SeaTurtleID2022} & \textbf{ZindiTurtleRecall} \\\midrule
 Sea turtle species & Loggerheads & Greens/Hawksbills \\
 Images & 8729 & 12803 \\
 Individuals & 438 & 2265 \\
 Image average size & 5269$\times$3564 & 1382$\times$1118 \\
 Head average size & 635$\times$554 & 1382$\times$1118 \\
 Location & underwater & land (rehab.\ centre) \\
 Allowed splits & \textit{time-aware} \& \textit{open-set} & \textit{random} \\
 \midrule
 In the wild          & \cmark~~~~~~ & \xmark~~~~~~ \\
 Turtle segment  & \cmark~~~~~~ & \xmark~~~~~~ \\
 Head bbox    & \cmark~~~~~~ & \cmark~~~~~~ \\ 
 Head segment    & \cmark~~~~~~ & \xmark~~~~~~ \\
 Head orientation     & \cmark~~~~~~ & partially \\
 Flipper segment & \cmark~~~~~~ & \xmark~~~~~~ \\
 Flipper bbox & \cmark~~~~~~ & \xmark~~~~~~ \\
 Timestamp & \cmark~~~~~~ & \xmark~~~~~~ \\
 \bottomrule
\end{tabular}
\caption{Comparison with the ZindiTurtleRecall dataset.
}
 \vspace{-1.0cm}
\label{table:dataset}
\end{table}

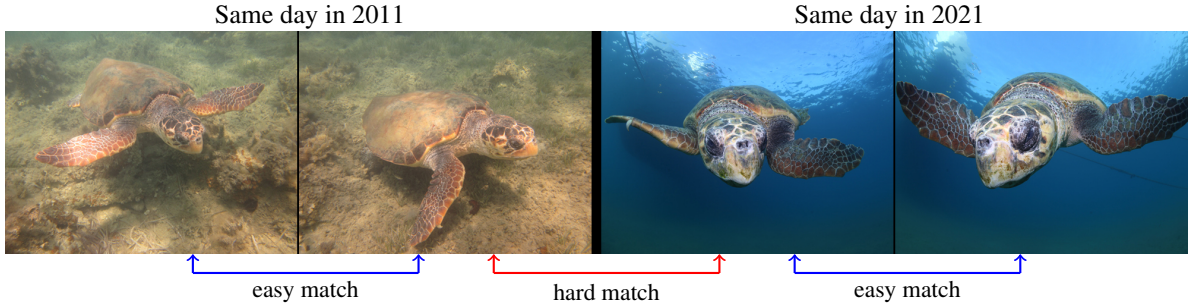
\begin{figure*}[!t]
\centering
Same day in 2011 \hspace{5cm} Same day in 2021\\
\begin{tikzpicture}
 \node[anchor=south west,inner sep=0]  at (0,0) {\includegraphics[width=0.9\linewidth]{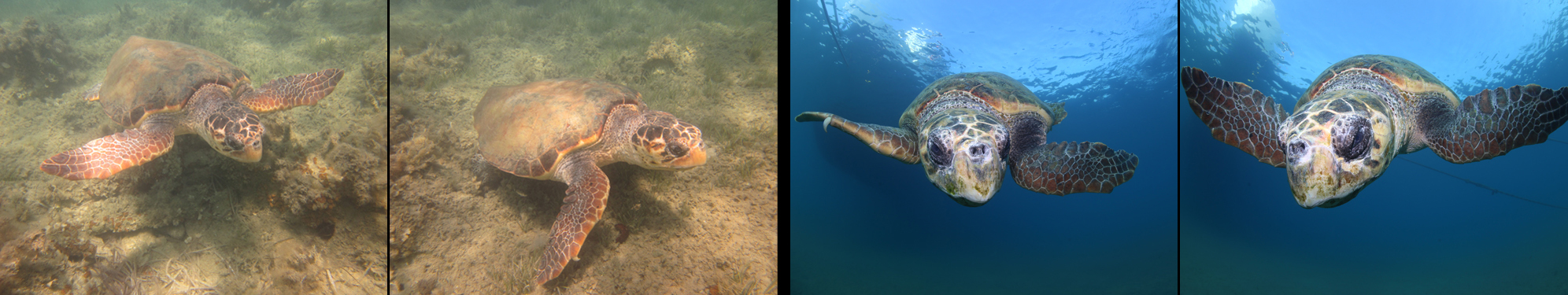}}; 
 \draw [thick,blue, <-] (2.5,0) to (2.5,-0.25);
 \draw[thick,blue]  (2.5,-0.25) to (5.5,-0.25);    
 \draw [thick,blue,->] (5.5,-0.25) to (5.5,0);
 \node at (4.0,-0.5) {\small{easy match}};
\draw [thick,blue, <-] (10.5,0) to (10.5,-0.25);
 \draw[thick,blue,]  (10.5,-0.25) to (13.5,-0.25);    
 \draw [thick,blue,->] (13.5,-0.25) to (13.5,0);
 \node at (12.0,-0.5) {\small{easy match}};

 \draw [thick,red, <-] (6.5,0) to (6.5,-0.25);
 \draw[thick,red]  (6.5,-0.25) to (9.5,-0.25);    
 \draw [thick,red,->] (9.5,-0.25) to (9.5,0);
  \node at (8.0,-0.5) {\small{hard match}};
\end{tikzpicture}
\vspace{-0.2cm}
\caption{Unwanted background similarities in photographs from same/similar locations or time of observations.}
\label{figure:time-aware}
\end{figure*}

\subsection{Dataset splits and subsets}\label{section:splits}

Standardly, the re-identification datasets are split into a reference (training) and a query set (test) randomly, which might result in unwanted data leakage and inflated performance. In other words, images from the same encounter might be in both sets. To illustrate the problem, we provide in Figure\,\,\ref{figure:time-aware} four images of the same individual turtle, two captured in the same day in 2011 and two in the same day in  2021. While images from the same day are easy to match due to the same background and coloration, images from different days/years do not share it and therefore are significantly more challenging to match. 
To overcome this issue, we introduce two realistic ecologically motivated splits that utilize timestamps to prevent information leakage from the test set to the training set. We further refer to those splits as time-aware splits. For easier future comparison, we provide predefined training/validation/test splits even though validation in some cases is not mandatory. The construction is further elaborated below. The dataset statistics, including the number of individuals and images, are listed in Table\,\,\ref{table:dataset_splits}. \vspace{-0.25cm} \\

\noindent\textbf{Time-aware closed-set split} is similar to a standard closed-set re-identification scenario, as all validation/test sets identities are available for training. Such a scenario is realistic in environments with controlled populations, e.g., zoos or reservations.
While constructing the split, we group all the data based on the date of acquisition and split it in a time-aware fashion. Data from approximately 80\% days are selected for the development set (training + validation), and the remaining days go to the test set. If an individual turtle was observed just once, it was kept for training. We provide 438 identities for training and 270 for testing. The development set was split into training/validation subsets using the same strategy.  \vspace{-0.25cm} \\

\noindent\textbf{Time-aware open-set split} is based on cutoff time points (specific years). In this setting, each subset (training/validation/test) contains all images within given subsequent periods. Intuitively, this split results in an open-set problem, reflecting the natural population dynamic and growth.
During construction, we used the 2010--2018 period for training, the whole year of 2019 for the validation, and the 2020--2022 period for the test set. There are 357 identities in the training set and 151 in the test set. Out of the 151 identities, 51 are newly observed. A similar ratio (\textit{new}/\textit{known}) is naturally acquired in the validation set; 38 out of 83 are new identities.
\begin{table}[!ht]
\small
\centering
\begin{tabular}{@{}lcccc@{}}
\toprule
& \multicolumn{2}{c}{\textit{\# of images }} & \multicolumn{2}{c}{\textit{\# of identities} } \\
\textbf{Subset}  & \textbf{closed-set} & \textbf{open-set} & \textbf{closed-set} & \textbf{open-set}  \\
\midrule
Training            & 4679 &  5303 & 438 &  357 \\
Validation          & 1418 &  1118 &~~91 & ~~83 \\
Test                & 2632 &  2308 & 270 &  151 \\
\bottomrule
\end{tabular}
\caption{Provided time-aware datasets split and their statistics.}
\label{table:dataset_splits}
\vspace{-0.25cm}
\end{table}

\noindent
\textbf{Note:} \textit{The open-set split is much closer to the real-world re-identification settings than the closed-set problem. Therefore, the open-set split should be preferred for automated method evaluation over all datasets.
In case closed-set evaluation is desired, then the time-aware split must be the preferred option over the random split.} \vspace{-0.15cm} \\

\noindent\textbf{Body-parts subsets:} Furthermore, we provide three subsets that cover various body parts, e.g., full-body, flippers, and heads, using crops from the original resolution. The number of data points differs for each body part, as some parts might not be visible. We used the time-aware closed-set and constructed part-based sets with the following number of training/test samples: (i) 6139 / 2650 full turtle bodies, (ii) 14849 / 6237 flippers, and (iii) 5956 / 2583 heads.

\section{Sea turtle re-identification baselines}\label{section:methods}

Animal re-identification is generally approached using either (i) traditional methods and local descriptors (e.g., SIFT and SURF)\,\cite{reno2019sift,andrew2016automatic,dunbar2021hotspotter}, (ii) deep learning\,\cite{bruslund2020re,ueno2022automatic,li2019atrw}, or (iii) species-specific methods\,\cite{bedetti2020system,weideman2020extracting,gilman2016computer}. To establish a baseline performance on the SeaTurtleID2022 and to propose the system for end-to-end turtle identification, we perform various ablation studies using various traditional and deep learning based methods. In this section, we describe selected methods and all relevant hyperparameters.

\subsection{Local feature-based methods}\label{section:feature}

The most popular methods used for wildlife re-identification -- Wild ID\,\cite{bolger2012computer}, and Hotspotter\,\cite{dunbar2021hotspotter} -- are based on local descriptors. Therefore we study the performance of SIFT and more recent Superpoint\cite{detone2018superpoint} descriptors on the proposed dataset. 
We have developed a straightforward algorithm (inspired by Dunbar et al.\,\cite{dunbar2021hotspotter}) based on local descriptors matching\footnote{For SIFT we use default parameters and OpenCV implementation; for Superpoint, we use default parameters and \href{https://github.com/magicleap/SuperPointPretrainedNetwork}{this implementation}.}.
First, we extract a set of keypoints and their corresponding descriptors for each image. 
Second, for all possible training-test image pairs, we calculate the distance between their descriptors.
Third, all potentially false matches are filtered out using a ratio test and threshold; the optimal values (0.2 for SIFT, 0.6 for Superpoint) for the ratio test thresholds were obtained from the training set.
At last, we predict an identity using the training label with maximal similarity score, calculated as an absolute number of correspondences.  
We opt not to use alternative approaches, such as RANSAC or SuperGlue, as they add significant computational overhead and provide just a small improvement\,\cite{dunbar2021hotspotter,pedersen2022re}.

\subsection{Metric learning}
Metric learning methods aim to learn a representation function that maps objects into a deep embedding space. Usually, a CNN- or transformer-based feature extractor is trained to group samples within the same semantic category closer and far from other categories. For our experiments, we use two algorithms with state-of-the-art performance in face recognition: ArcFace\,\cite{deng2019arcface} and Triplet loss\,\cite{schroff2015facenet}. For baseline performance evaluation, we use a Swin-B\,\cite{liu2021swin} backbone and default training hyperparameters.

Both metric learning approaches were optimized for 100 epochs using a learning rate of 0.01, the cosine annealing schedule, and a mini-batch size of 128. All images were pre-processed using the Random augment method.

\textbf{ArcFace loss}\,\cite{deng2019arcface} was designed for face recognition but can be easily repurposed for wildlife re-identification. It extends the cross-entropy loss by placing the embeddings on the hypersphere with radius $s$ and incorporating an angular margin $m$ to improve the learned embeddings' discriminative capability that ensures high inter-class variety while keeping a high level of intra-class compactness. The similarity of samples is determined using cosine distance. We use the same values for $s=64$ and $m=0.5$ as in\,\cite{deng2019arcface}.

In \textbf{Triplet loss}\,\cite{schroff2015facenet}, we select triplets $(x_a, x_p, x_n)$ with anchor $x_a$ that has the same label as positive $x_p$ and a different label than negative sample $x_n$. Triplet loss learns a representation that minimizes the distance between $x_a$ and $x_p$ and maximizes the distance between $x_n$ up to a margin $m$. In our experiments, we follow\,\cite{schroff2015facenet} and use $m=0.1$. Triplet loss tends to be sensitive to triplet selection. Therefore, we follow\cite{hermans2017defense} and select hard triplets using an online mining strategy to improve the training. \vspace{-0.15cm} \\

\noindent\textbf{Feature matching}: In our metric learning experiments, we approach animal identification using k-NN classifier in a deep embedding space. For each image $x$ from the test set, we assume its $k$ most similar training set identities, and we take the one with the highest occurrence. 
The formal definition of k-NN we use is as follows. The set of $k$ nearest neighbors  $S_x$ of $x$ is defined as a subset of the training set 
such that every point in the training set but not in $S_x$ is at least as far away from $x$ as the furthest point in $S_x$, measured in a suitable distance function.
We define the classifier as a function returning the most common label in $S_x$
In the case of a draw, we take an identity from smaller $k$, i.e, $(k-1)$. For metric learning approaches, the distance function is a cosine distance, i.e.,
\begin{equation}
\text{dist}(x,z) = \frac{x \cdot z}{\|x\|\ \|z\|}.
\end{equation}

\subsection{Random vs. time-aware splits}
To showcase the unwanted performance overestimation when a random dataset split is used, we compare the performance of newly proposed time-aware splits (open and closed) with their random counterparts. 
The random split is obtained by randomly shuffling the time-aware split for each identity separately. This ensures a fair comparison between the split with the same training/validation/test ratio. We used the entire image and different body parts in this experiment. We use an ArcFace loss with the Swin-B backbone and input size of $224\times224$.

\section{Baseline Results}
In this section, we provide 
(i) baseline results for body-part segmentation and re-identification achieved over the newly proposed dataset
(ii) qualitative and quantitative evaluation to show the importance of the time-aware splits,
and (iii) performed ablation studies to select the most viable approach for sea turtle re-identification. 

Based on extensive experiments with different $k$ values for k-NN matching (available in Supplementary), we predict an identity using k-NN, with $k = 1$. \vspace{-0.15cm} \\

\noindent\textbf{Local vs deep features}:
Comparing local descriptors with metric learning approaches showed superior performance of metric learning on our dataset and seven other datasets with patterned species. In most cases, the metric learning approaches outperformed the Superpoints by more than 20\%. Furthermore, if we compare local descriptor methods, the Superpoints method is a better fit for animal re-identification. A detailed comparison is listed in Table\,\,\ref{table:split_closed}. \\

\begin{table}[!h]
\setlength{\tabcolsep}{0.35em} 
\small
\centering
\begin{tabular}{@{}lcccc@{}}
\toprule
\textbf{Dataset} & \textbf{SIFT} & \textbf{Superpoint} & \textbf{ArcFace} & \textbf{Triplet} \\
\midrule
BelugaID\,\cite{belugaid}                   &~\,1.1 &~\,2.4 & \colorbox{LightGreen}{18.2} & \colorbox{Green}{20.5} \\
HumpbackWhaleID\,\cite{humpbackwhale}       &  11.7 &  11.8 & \colorbox{Green}{52.5} & \colorbox{LightGreen}{43.9} \\
NDD20\,\cite{trotter2020ndd20}              &  17.1 &  \colorbox{LightGreen}{30.0}  & \colorbox{Green}{59.1} & 29.9 \\
NOAARightWhale\,\cite{rightwhale}           &~\,6.5 &  15.3 & \colorbox{Green}{23.5} & 5.4 \\
WhaleSharkID\,\cite{holmberg2009estimating} &~\,4.3 &  22.9 & \colorbox{LightGreen}{28.6} & \colorbox{Green}{32.5} \\
\midrule
ZindiTurtleRecall\,\cite{zinditurtles}      &  17.9 &  \colorbox{LightGreen}{25.7}  & \colorbox{Green}{45.8} & 19.1 \\
SeaTurtleID2022 (ours)                           &~\,8.4 & 20.2 & \colorbox{Green}{34.7} & \colorbox{LightGreen}{25.7} \\
\bottomrule
\end{tabular}
\caption{Local and deep feature methods performance comparison (accuracy) for full images. Time-aware closed-set split. Input size $224\times224$. For metric learning, a Swin-B backbone was used.}
\vspace{-0.5cm}
\label{table:split_closed}
\end{table}

\noindent\textbf{Body parts performance:} In addition to setting overall full-body turtle performance, we explored the importance of various body parts, revealing their relative significance. 
In contrast to the findings of\,\cite{mills2023photo}, our results highlight the key role of the turtle's \textit{head} in sea turtle identification.
Focusing solely on the \textit{head} increased the absolute performance by 34.5\% compared to the full body. Furthermore, we show that the \textit{flippers} appear as the less influential body part for in-the-wild identification using metric learning\footnote{For the flippers performance evaluation, we choose the closest (based on cosine similarity) identity using all available flippers on a given image.}. The full comparison is provided in Table\,\,\ref{table:body_parts}. \vspace{-0.15cm} \\

\noindent\textbf{Encounter based prediction}:
Available timestamps allow combining all image-based predictions into so-called encounters. Rather than identifying each image separately, one identity is predicted for each set of images belonging to one individual. Using just majority voting to combine the image-based predictions, we significantly increased the performance for all body parts (see Table\,\,\ref{table:body_parts}). In the case of heads, the accuracy was increased by 19.2\%.

\subsection{Random vs time-aware splits}
The performance comparison of two ArcFace-trained feature extractors on the random and time-aware splits of the SeaTurtleID2022 dataset validated our hypothesis about unwanted performance inflation related to training-to-test data leakage. Results listed in Table\,\,\ref{table:body_parts} demonstrate that the random split results (in terms of accuracy) were higher by 42.2\%, 53.8\%, 45.8\%, and 18\% for full image, and flippers, body, and head crops, respectively.

\begin{table}[!h]
\setlength{\tabcolsep}{0.475em} 
\small
\centering
\begin{tabular}{@{}rlrrrr@{}}
\toprule
& \textbf{Split}      & \textbf{Full image} & \textbf{Flippers} & \textbf{Turtle} & \textbf{Head}  \\
\midrule
\scriptsize{Images} \hspace{-0.3cm} & Time-aware   & 17.1 &  12.2 & 34.7 & 69.2 \\
\scriptsize{Encounters} \hspace{-0.3cm} & Time-aware         & -- & 21.4 & 48.6 & 88.4 \\
\midrule
\scriptsize{Images} \hspace{-0.3cm} & Random        & \textit{59.4} & \textit{66.0} &\textit{80.5} & \textit{87.2} \\
\bottomrule
\end{tabular}
\caption{Random split accuracy inflation on  SeaTurtleID2022 (closed-set). Encounter- vs image-based; Swin-B + ArcFace.}
\label{table:body_parts}
\end{table}

\noindent\textbf{Performance inflation analysis}:
To further elaborate on the performance inflation, we conducted an additional re-identification experiment using (i) images with redacted backgrounds, showing only the turtle in the foreground, and (ii) images with redacted foregrounds, displaying only the background.
With the redacted background, the model's performance remains relatively comparable to the full image performance in both scenarios. Contrarily, in the case of redacted foreground, the model trained on a random split exhibits comparable performance to that achieved on the full images. However, the performance for the model trained on a time-aware dropped significantly in performance relative to the full images, achieving only 3.9\% accuracy. See results in Table\,\,\ref{table:perf_inflation}.

\begin{table}[!h]
\small
\centering
\begin{tabular}{lrrr}
\toprule
\textbf{Split} & \textbf{Full image} & \textbf{Background} & \textbf{Foreground} \\
\midrule
Random      & 59.4 & 45.1 & 59.5 \\
Time-aware  & 17.1 &3.9 & 14.3 \\
\midrule
~~~~~~$\Delta$  & +\textit{42.2} & +\textit{41.2} & +\textit{45.2} \\
\bottomrule
\end{tabular}
\caption{Random split accuracy inflation on the SeaTurtleID2022 (closed-set). Swin-B + ArcFace; $224\times224$.}
\label{table:perf_inflation}
\end{table}

\begin{figure}[!ht]
 \centering
  \includegraphics[width=0.475\linewidth, height=3cm]{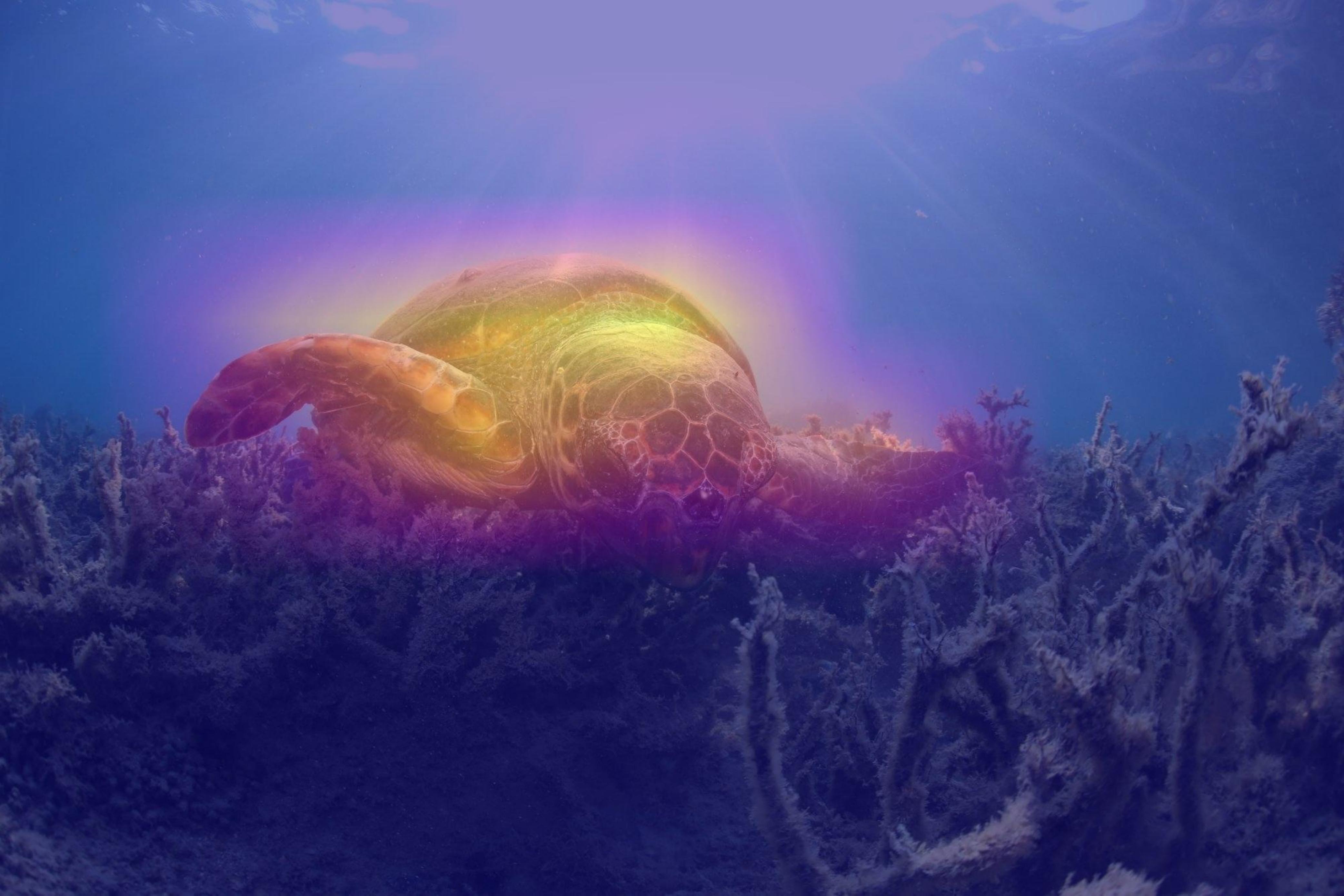}
  \includegraphics[width=0.475\linewidth, height=3cm]{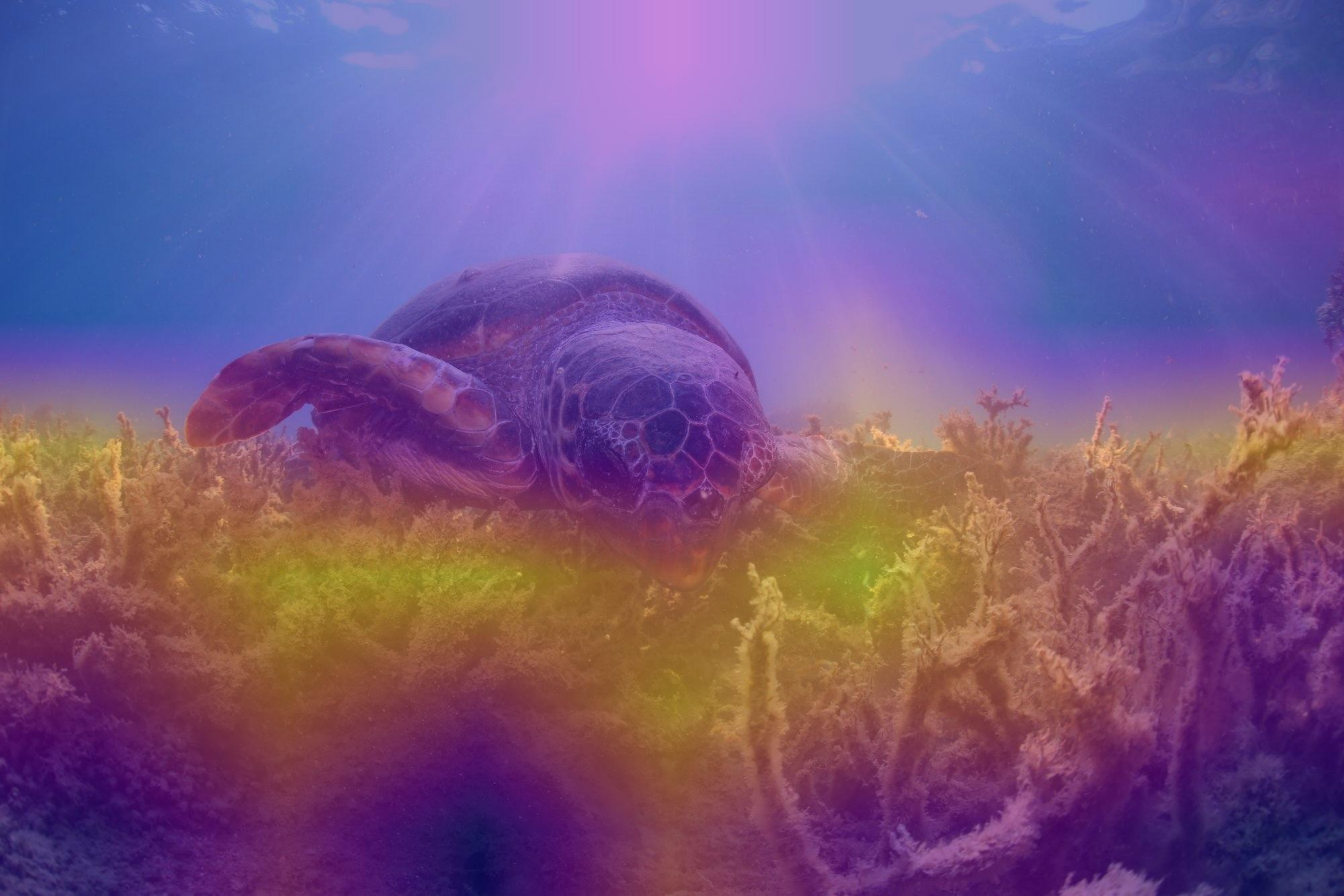} \\
 \vspace{0.05cm}
  \includegraphics[width=0.475\linewidth, height=3cm]
  {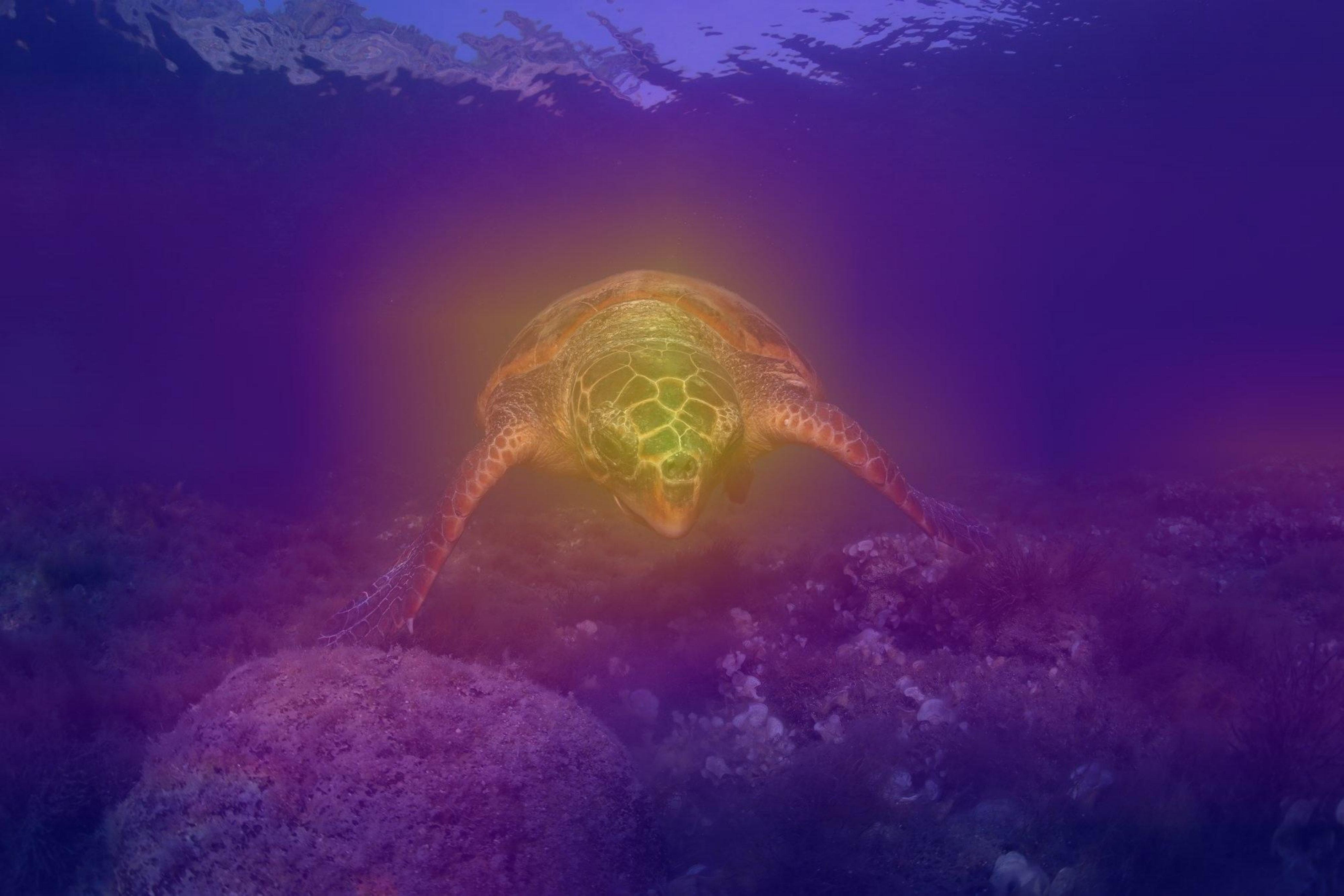}
  \includegraphics[width=0.475\linewidth, height=3cm]
  {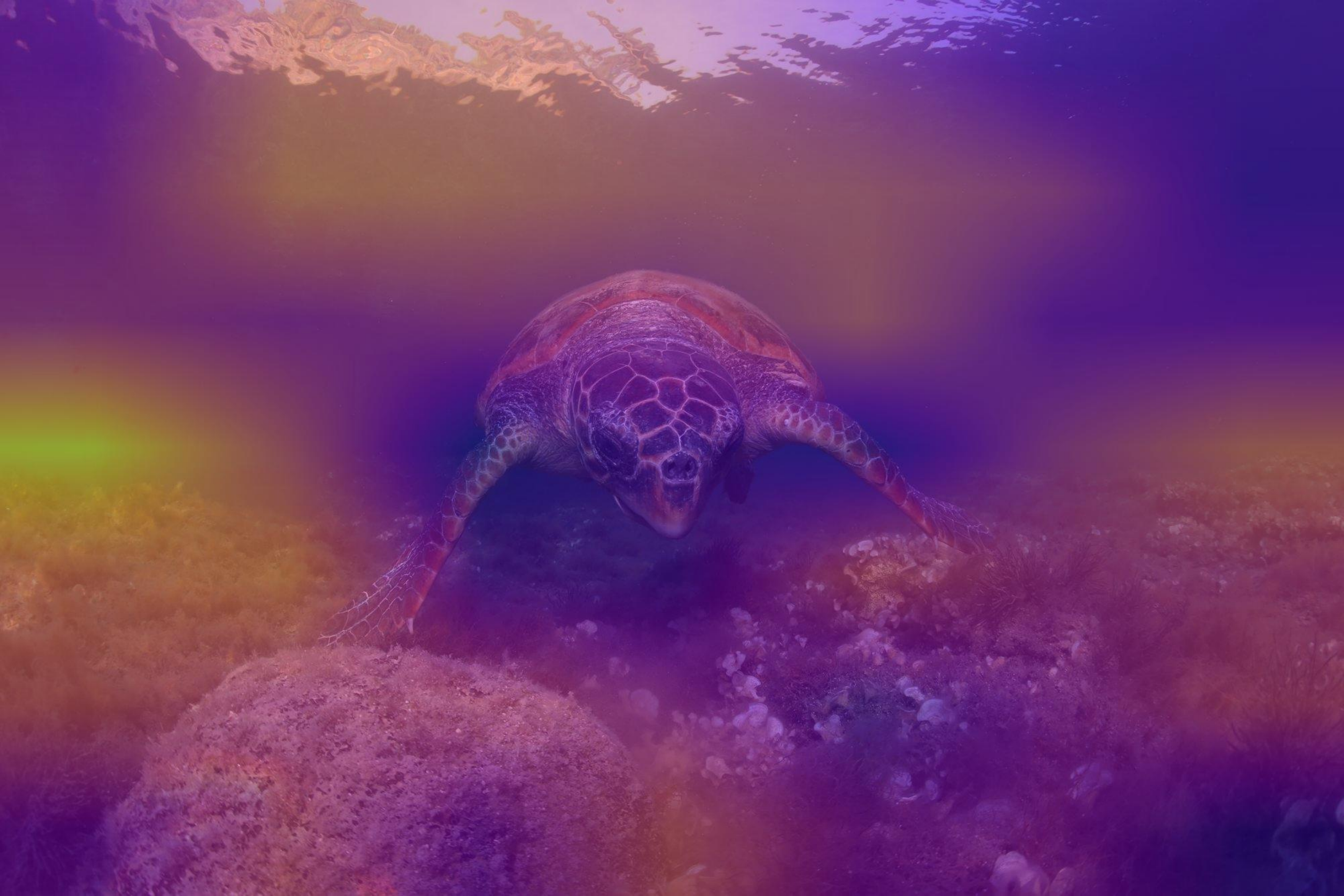}
 \caption{
 Qualitative evaluation demonstrating overfitting to the background on random split using Grad-CAM++. 
 Identity-based activations for (left) time-aware and (right) random split.
 }
 \label{fig:gradcam}
 
\end{figure}

Furthermore, we qualitatively demonstrate overfitting to the background using Grad-CAM++\,\cite{chattopadhay2018grad} and visualizing identity activations based on the cosine similarity between the embeddings of the two images. We selected two similar images with noticeable backgrounds from the same encounter that are in the test set for both random and time-aware splits. In Figure\,\,\ref{fig:gradcam}, we illustrate that the model trained on the random split learns to utilize background features, whereas the model trained using the time-aware approach concentrates on the turtle's features.

\newpage

\subsection{Body-parts segmentation baselines}\label{section:methods}
The SeaTurtleID2022 dataset comes along with instance segmentation annotations; thus, it might be used as a benchmark for instance segmentation or object detection. To set the baseline performance for the turtle body parts (head, flipper, and full-body) segmentation, we evaluate three distinct architectures, including the standard Mask R-CNN\,\cite{He_2017_ICCV}, the Hybrid Task Cascade (HTC)\,\cite{chen2019hybrid}, and the state-of-the-art transformer-based Mask2Former\,\cite{cheng2021per}. 
We combine the three detection methods with two backbones, ResNet-50\,\cite{he2016deep} and Swin-B transformer\,\cite{liu2021swin} using the MMDetection\,\cite{mmdetection} framework. While training, both backbones were initialized from publicly available ImageNet-1k weights using the default implementation and hyperparameters setting. All models were fine-tuned for 12 epochs with a step-wise learning rate schedule. Experiments are conducted on both time-aware splits.

Generally, all selected methods evaluated on the SeaTurtleID2022 achieved a competitive performance (in terms of coco mAP) suitable for the following task, i.e., turtle re-identification. While the best-performing model --\textit{ Mask2Former with Swin-B backbone} -- achieved a coco mAP of 0.896, the worst-performing model  -- \textit{Mask R-CNN with ResNet-50 backbone} -- achieved an mAP of 0.865. Even though the Mask2Former approach showed better overall performance, the HTC method performed better on heads that are important for accurate re-identification.

The full performance comparison on both time-aware splits (open and closed) is available in Table\,\,\ref{table:instance_segmentation_closed} and supplementary materials.

\begin{table}[!hb]
\small  
\centering
\setlength{\tabcolsep}{0.475em} 
\begin{tabular}{@{}llc|ccc@{}}
\toprule
& \textbf{Method}                     & mAP & ${head}$ & ${turtle}$ & ${flippers}$ \\
\midrule
\parbox[t]{-2mm}{\multirow{3}{*}{\rotatebox[origin=c]{90}{\scriptsize{ResNet-50}}}} \hspace{-0.25cm}  
& Mask R-CNN  & 0.865 & 0.838 & 0.910 & 0.848 \\
& HTC         & 0.868 & 0.842 & 0.912 & 0.852 \\
& Mask2Former & \colorbox{LightGreen}{0.892} & 0.822 & \colorbox{Green}{0.977} & \colorbox{LightGreen}{0.876}\\
\midrule
\multirow{3}{*}{\rotatebox[origin=c]{90}{\scriptsize{Swin-B}}} \hspace{-0.25cm}                      
& Mask R-CNN  & 0.871 & \colorbox{LightGreen}{0.845} & 0.919 & 0.85 \\
& HTC         & 0.880 & \colorbox{Green}{0.860} & 0.923 & 0.857 \\
& Mask2Former & \colorbox{Green}{0.896} & 0.829 & \colorbox{LightGreen}{0.975} & \colorbox{Green}{0.883} \\ 
\bottomrule
\end{tabular}
\caption{Instance segmentation performance of selected \textit{backbone} and \textit{head} architectures over the SeaTurtleID2022. Closed-set split.}
\label{table:instance_segmentation_closed}
\vspace{-0.5cm}
\end{table}

\section{Recommended end-to-end system}
Following the insights from our baseline experiments allowed us to create a reliable end-to-end system that takes sets of images as input and returns identity predictions. The system within the pipeline and the performance of the system are fully described below.

\textbf{First}, we find all head region bounding boxes on high-resolution images (20MP) using the Hybrid Task Cascade instance segmentation model (with Swin-S backbone). We focus primarily on turtle heads as they allow the best re-identification capability.
\textbf{Second}, we crop all heads from the high-resolution photographs and rescale them to $224\times224$ to match the expected input size for the feature extractor, i.e., the Swin-B ArcFace-trained re-identification model.
\textbf{Third}, all head-based crops are feed-forwarded into the feature extractor to obtain feature vectors for matching. For images without a head segmentation, we do not provide any identity prediction.
\textbf{Fourth}, for each image, we predict an identity using \mbox{k-NN} (with k = 1) with the training set's head embeddings. 
\textbf{Finally}, we group all images based on time and create the so-called encounters. The identity of each image within an encounter is retrieved by majority voting.

The proposed end-to-end system for sea turtle re-identification achieved an accuracy of 86.8\% on the SeaTurtleID2022. Notably, it shows a significant improvement over the 17.1\% accuracy achieved by a naive approach that analyzes full images without utilizing body parts or harnessing encounter knowledge.

\section{Conclusions}
This paper introduced the \href{https://www.kaggle.com/datasets/wildlifedatasets/seaturtleid2022}{SeaTurtleID2022 dataset}, the longest-spanned publicly available wildlife re-identification dataset with various annotations, e.g., identities, encounter timestamps, and body parts segmentation masks. The dataset can be used for benchmarking re-identification algorithms and several other computer vision tasks, including instance and semantic segmentation and object detection. Instead of a standard "\textit{random}" split, we highlight the necessity to use  realistic and ecologically motivated splits: 
(i) \textit{time-aware}: with reference  data from different encounters, and 
(ii) \textit{open-set}: with new \textit{unknown} individuals (i.e., newly introduced to population) in test and validation sets.

Furthermore, (i) we  provided a baseline performance of various methods, for instance segmentation and animal re-identification,
(ii) provided qualitative and quantitative evidence that time-unaware (random) splits of the dataset lead to a significant performance overestimation bias,
and (iii) we proposed, described, and evaluated an end-to-end system for sea turtle identification in the wild, that could potentially be transferred to other species as well.

\section{Acknowledgments}
This research was supported by the Czech Science Foundation (GA CR), project No. GA22-32620S and by the Technology Agency of the Czech Republic, project No. SS05010008.
Computational resources were provided by the e-INFRA CZ project (ID:90254), supported by the Ministry of Education, Youth and Sports of the Czech Republic and by the OP VVV project ``Research Center for Informatics'' (No. CZ.02.1.01/0.0/0.0/16\_019/0000765).


{\small
\bibliographystyle{ieee_fullname}
\bibliography{bibliography,bibliography2}
}

\appendix

\section{Additional Experiments and Evaluation}

\subsection{Dependence of  identification accuracy  on \textit{k} \mbox{in \textit{k}-NN classification}}

We conducted additional experiments to find an optimal \textit{k} value for the animal re-identification using the \textit{k}-NN classifier. Besides  SeaTurtleID2022 (head and full-body versions), we evaluated the experiments on BelugaID, NDD20, WhaleSharkID, HumpbackWhaleID, NOAARightWhale and ZindiTurtleRecall datasets. We used embeddings from the ArcFace-trained model.

\begin{figure}[!ht]
\centering
\includegraphics[width=0.95\linewidth]{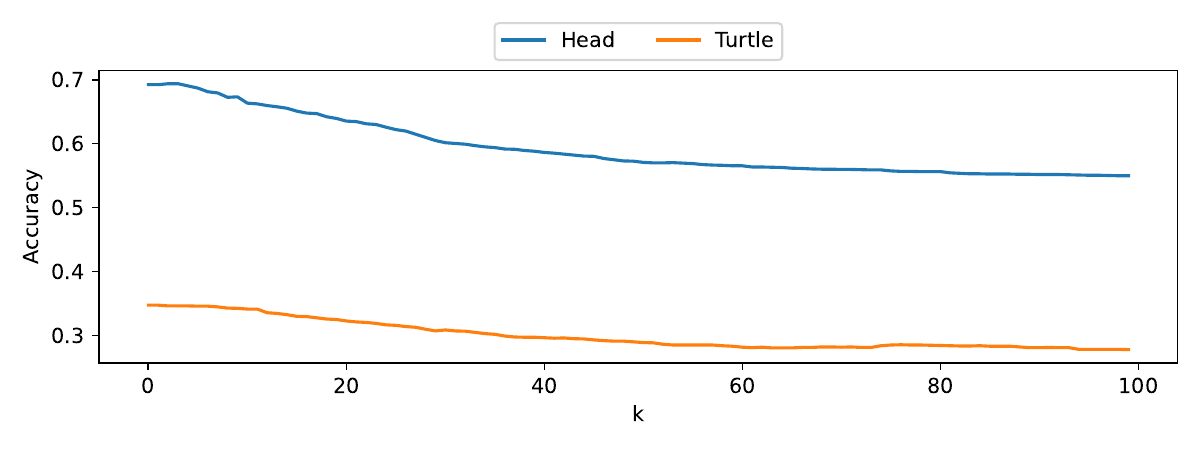}
\includegraphics[width=0.95\linewidth]{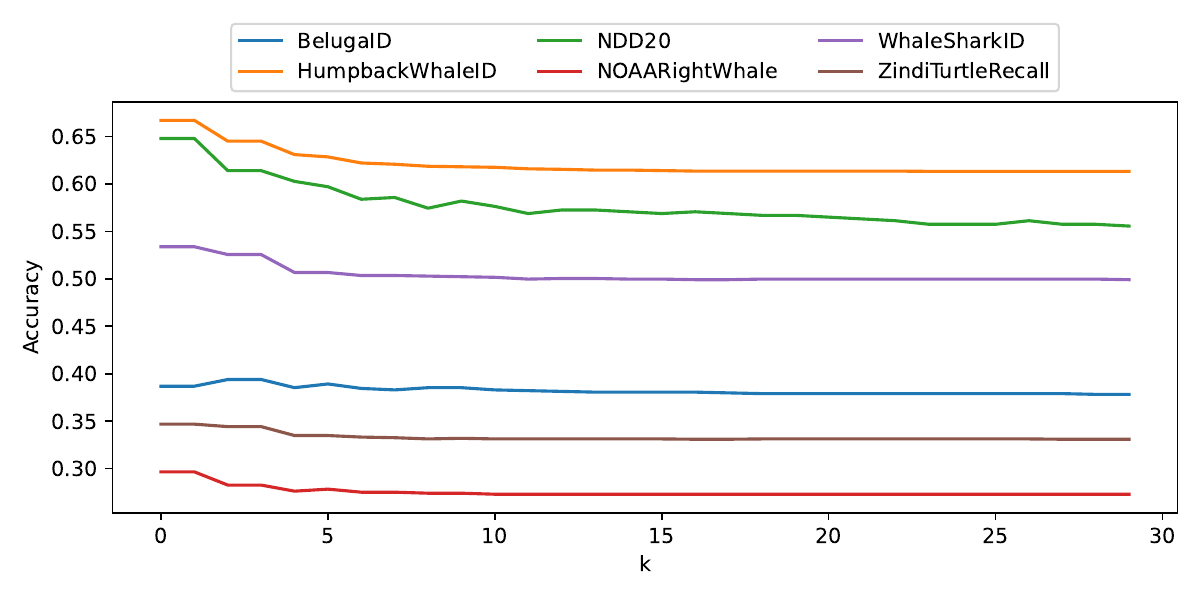}
\caption{\textbf{Effect of \textit{k} on performance}. We display the classification accuracy of \textit{k}-NN classifier with ArcFace embeddings for various \textit{k} values. Different body parts (e.g.\ head and full-body) performance on the SeaTurtleID2022 dataset (top) and selected wildlife re-identification datasets (bottom).}
\label{fig:knn-experiment}
\end{figure}

Our findings indicate that opting for a smaller \textit{k} value yields better results, with \textit{k}=$1$ being a reasonable choice in any case. This discovery is consistently supported by results in various other datasets we considered. We attribute this phenomenon to the significant class imbalance present in wildlife datasets. As \textit{k} increases, identities with higher prior probability overwhelm the classification results, i.e., for larger \textit{k} values, there are often just a few samples for the less frequent identities.
On the SeaTurtleID2022 dataset (head and full-body) the performance in terms of accuracy significantly decreased from 69.2\% at $\textit{k}=1$ to 55.0 \% at $\textit{k}=100$. A similar, though less severe, drop in performance was also noticeable in other datasets. We depict the relationship between accuracy and values of $k$ in  \cref{fig:knn-experiment}.

\subsection{Time-aware vs random split: Additional experiment with cross-entropy learning}

To further elaborate the performance inflation related to random split, we have tested various deep learning backbone architectures optimized using softmax cross-entropy. In \cref{table:different_backbones}, we provide the performance of five architectures on two splits of the SeaTurtleID2022 dataset: time-aware and a random split. In \cref{table:different_datasets}, we perform a similar experiment on 3 other datasets that allow time-aware splitting, showcasing that this inflation is not a characteristic of the SeaTurtleID2022 dataset, but it occurs in other datasets as well. We employed a 50/50 training-test split; therefore, results are directly not comparable to results in Section 4.1. In all experiments, all images were resized to match the pre-trained model input size of $224\times224$. 

\begin{table}[!h]
\centering
\begin{tabular}{@{}lrr@{}}
\toprule
 \textbf{Backbone} & \textbf{Time-aware close-set} & \textit{Random split} \\
\midrule 
 ResNeXt-50       & 38.6\% & \textit{63.4\%} \\
 EfficientNet-B0  & 39.9\% & \textit{76.5\%} \\
 ConvNeXt-B       & 47.2\% & \textit{78.5\%} \\
 ViT-Base/p32     & 45.2\% & \textit{82.5\%} \\
 Swin-B/p4w7      & 47.6\% & \textit{83.2\%} \\ 
\bottomrule 
\end{tabular}
\caption{Performance inflation (accuracy) with different backbones fine-tuned with softmax cross-entropy.}
\label{table:different_backbones}
\end{table}

\begin{table}[!h]
\centering
\begin{tabular}{@{}lrr@{}}
\toprule
\textbf{Dataset} & \textbf{Time-aware close-set} & \textit{Random split} \\
\midrule 
BelugaID            & 7.8\% & \textit{12.1\%}  \\
GiraffeZebraID      & 2.1\% & \textit{30.1\%}  \\
MacaqueFaces        & 91.1\% & \textit{98.9\%} \\
\bottomrule 
\end{tabular}
\caption{Performance inflation (accuracy) with different datasets.}
\label{table:different_datasets}\vspace{-0.35cm}
\end{table}

\subsection{Body-part instance segmentation}

We further present additional baseline instance segmentation experiments for different turtle body parts (head, flipper, and full-body). 
We provide an evaluation of three architectures, e.g., Mask R-CNN, the Hybrid Task Cascade (HTC), and the state-of-the-art transformer-based Mask2Former, on the time-aware open-set split. We used the same training strategy, i.e., backbones were initialized from publicly available ImageNet-1k checkpoints using the default implementation and hyperparameters setting and fine-tuned the models for 12 epochs with a step-wise LR schedule.

The comparison of selected methods utilizing well-known CNN- and transformer-based backbone architectures on the time-aware open-set split of the SeaTurtleID2022 dataset validated findings from the initial experiment with closed-set split, i.e., that the Mask2Former (both backbones) approach showed better overall performance but underperformed in the on heads. See Table\,\,\ref{table:instance_segmentation_open} for detailed performance evaluation.

\begin{table}[!h]
\small  
\centering
\setlength{\tabcolsep}{0.475em} 
\begin{tabular}{@{}llc|ccc@{}}
\toprule
& \textbf{Method}                     & mAP & ${head}$ & ${turtle}$ & ${flippers}$ \\
\midrule
\parbox[t]{-2mm}{\multirow{3}{*}{\rotatebox[origin=c]{90}{\scriptsize{ResNet-50}}}} \hspace{-0.25cm}  
& Mask R-CNN  & 0.827 & 0.735 & 0.907 & 0.840 \\
& HTC         & 0.833 & \colorbox{LightGreen}{0.740} & 0.909 & 0.849 \\
& Mask2Former & \colorbox{LightGreen}{0.850} & 0.708 & \colorbox{LightGreen}{0.975} & \colorbox{LightGreen}{0.866}\\
\midrule
\multirow{3}{*}{\rotatebox[origin=c]{90}{\scriptsize{Swin-B}}} \hspace{-0.25cm}                      
& Mask R-CNN  & 0.833 & \colorbox{Green}{0.743} & 0.913 & 0.844 \\
& HTC         & 0.839 & \colorbox{LightGreen}{0.740} & 0.921 & 0.856 \\
& Mask2Former & \colorbox{Green}{0.855} & 0.714 & \colorbox{Green}{0.977} & \colorbox{Green}{0.874} \\ 
\bottomrule
\end{tabular}
\caption{Instance segmentation performance of selected \textit{backbone} and \textit{head} architectures over the SeaTurtleID. Open set split.}
\label{table:instance_segmentation_open}
\vspace{-0.25cm}
\end{table}

\subsection{The importance of time-aware splitting}
\begin{figure}[!h]
\vspace{-0.5cm}
\begin{tikzpicture}[font=\small]
\begin{axis}[
      width=\linewidth,
      ymin=0,
      ylabel={correct match proportion},
      xtick={1,2,3,4,5,6},
      xticklabels={same day, 1 day, 1 week, 1 month, 1 year, more\phantom{1}},
      ytick={0.1,0.2,0.3,0.4,0.5},
      yticklabels={0.1,0.2,0.3,0.4,0.5},
      xtick pos=left,
      height=130,
      ybar,
      bar width=20pt,
]
      \addplot table[x index=0, y index=1] {\dataMatchesTime}; 
\end{axis}
\end{tikzpicture}
\caption{The probability of the correctly matched pairs of images of the same individuals with the same head orientation (left side) decreases as the time between the two images increases.}
\label{fig:matches_time}
\end{figure}
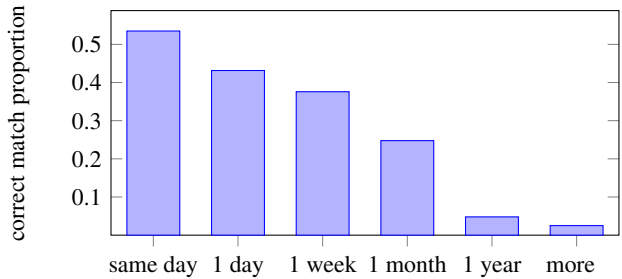

We further test and demonstrate the need for time-aware splits on other datasets that include timestamps using the Swin-B/p4w7 with the same setting as in the previous section. In \cref{table:different_datasets}, we show that in all cases, the results on the random split are undesirably inflated and much better than the ones of the time-aware split.
We get further insight by considering all pairs of images of the same individuals with the same head orientation and see how their matching probability (proportion of correctly matched pairs) is affected by the time between them. \cref{fig:matches_time} shows that the probability of correctly matching such image pairs decreases as the time between them increases. For instance, while this probability is 53.5\% for images taken on the same day, it decreases to 2.5\% for images taken more than one year apart. \\

\noindent\textbf{Further insights}: We further interpret \cref{fig:matches_time} with the specific example of turtle ``t298'', which was observed only on two days: 01/07/2016 and 12/07/2020. The random split has images from both dates in both reference and query sets, while the time-proportion split contains all images from 2016 in the reference set and all images from 2020 in the query set. While there were 26 matches for 2016-2016 images and 140 matches for 2020-2020 images, there were only 2 matches for 2016-2020 images. This further implies that there are many matches between the reference and query sets for the random split but almost no such matches for the time-proportion split. Therefore, the random split unnaturally simplifies the real-world re-identification problem. \\

\section{Additional figures about the SeaTurtleID2022 dataset}

\begin{figure}[!ht]
\centering
\begin{minipage}[t]{0.32\linewidth}
\centering
\includegraphics[width=0.95\textwidth]{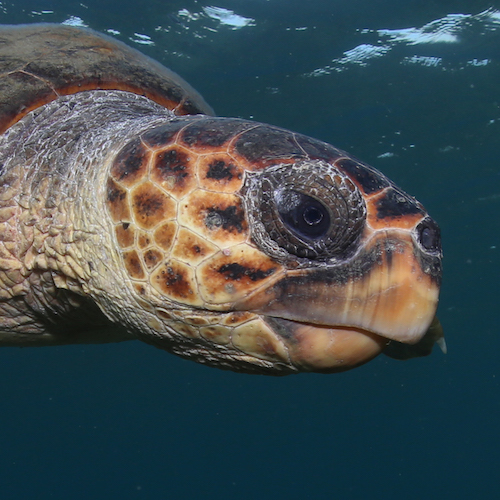}\\[0.5em]
\includegraphics[width=0.95\textwidth]{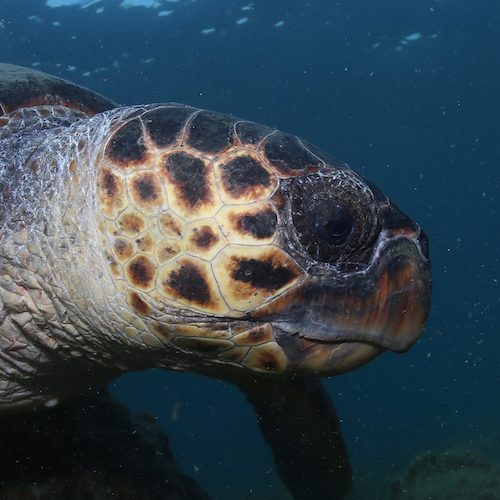}\\[0.5em]
\includegraphics[width=0.95\textwidth]{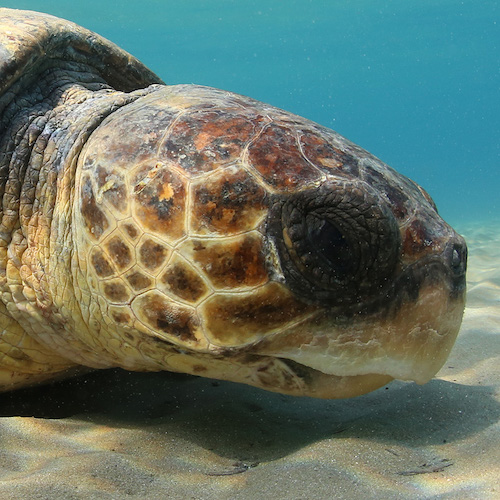}\\[0.5em]
\includegraphics[width=0.95\textwidth]{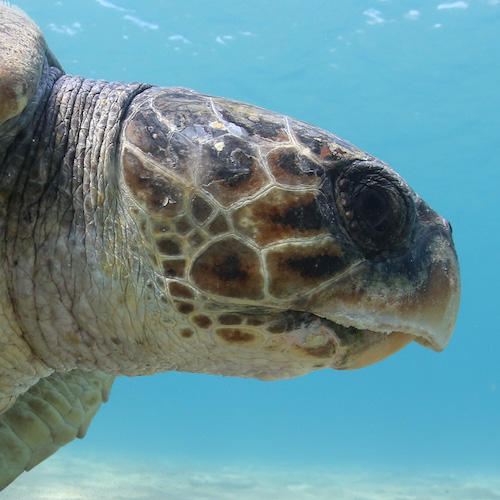}\\[0.5em]
\includegraphics[width=0.95\textwidth]{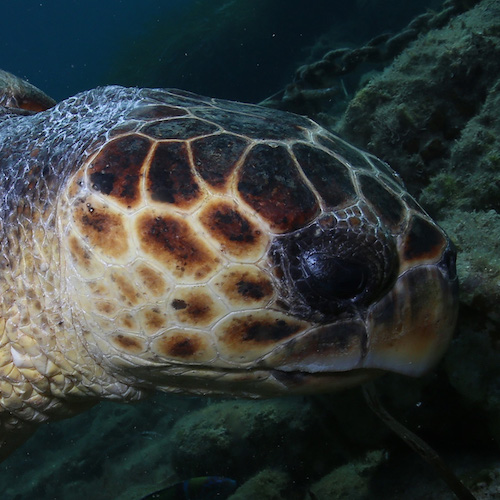}\\[0.5em]
\includegraphics[width=0.95\textwidth]{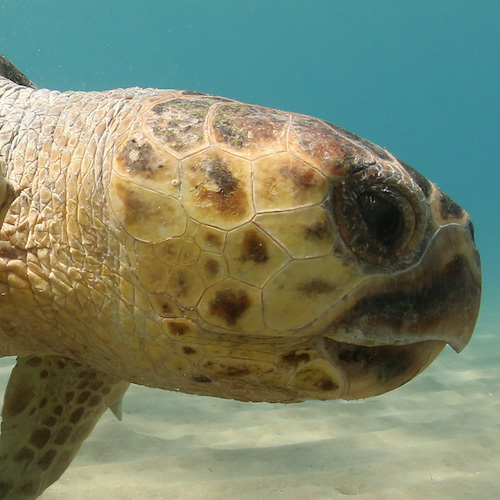}\\[0.5em]
\includegraphics[width=0.95\textwidth]{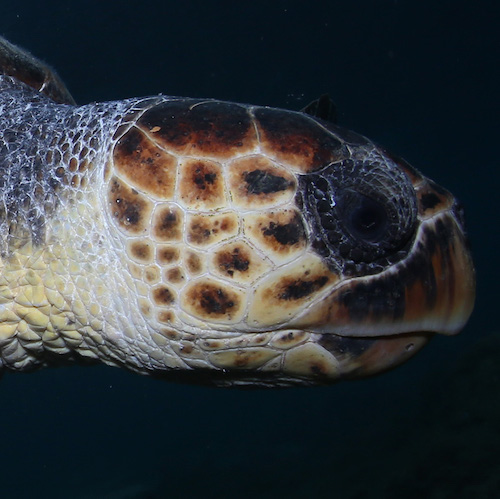}\\[0.5em]
\end{minipage}
\begin{minipage}[t]{0.32\linewidth}
\centering
\includegraphics[width=0.95\textwidth]{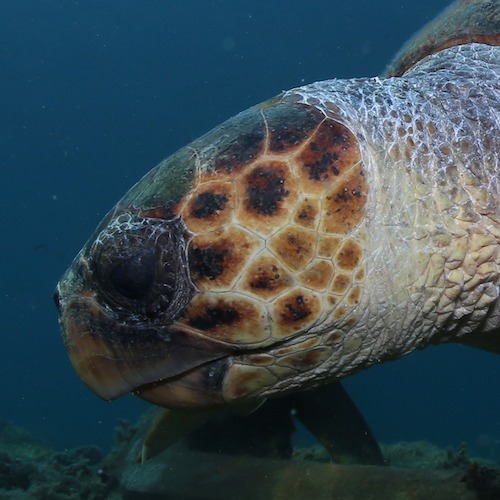}\\[0.5em]
\includegraphics[width=0.95\textwidth]{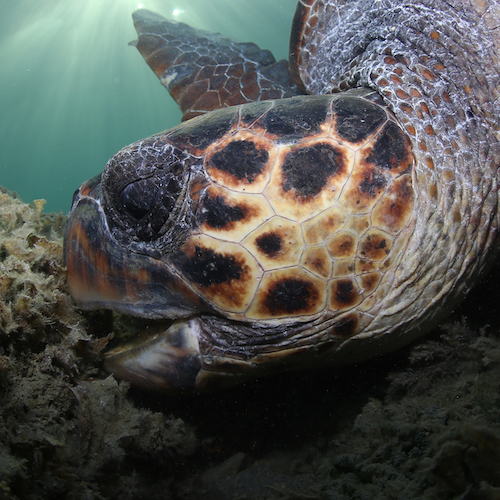}\\[0.5em]
\includegraphics[width=0.95\textwidth]{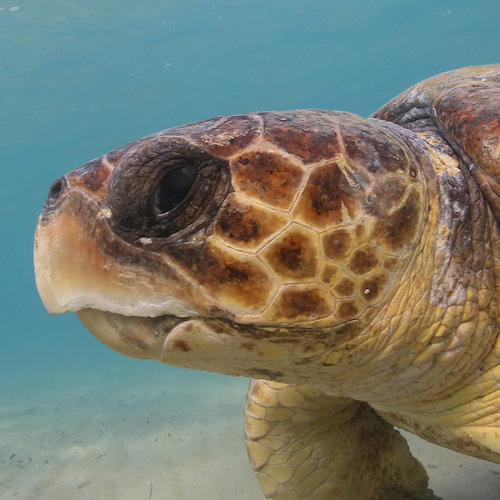}\\[0.5em]
\includegraphics[width=0.95\textwidth]{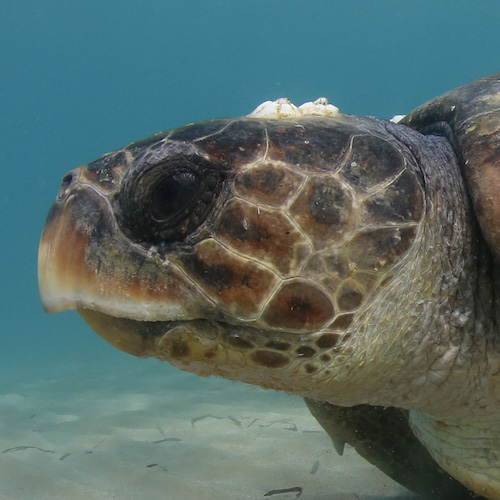}\\[0.5em]
\includegraphics[width=0.95\textwidth]{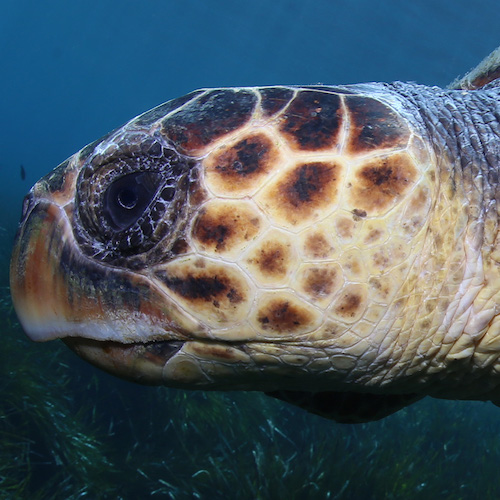}\\[0.5em]
\includegraphics[width=0.95\textwidth]{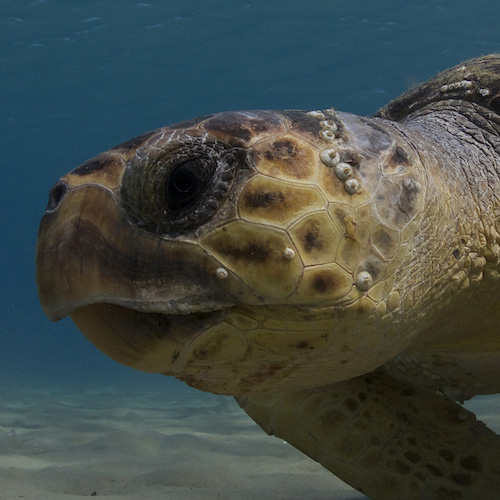}\\[0.5em]
\includegraphics[width=0.95\textwidth]{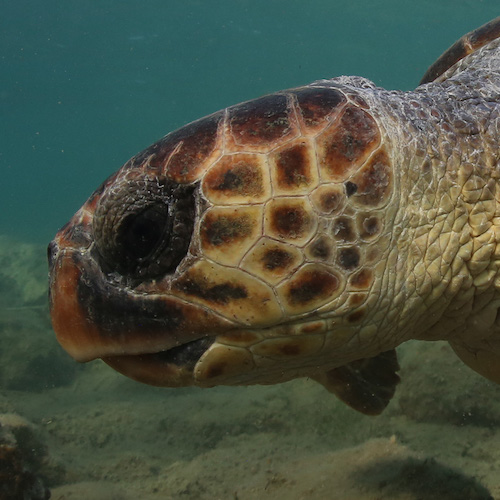}\\[0.5em]
\end{minipage}
\begin{minipage}[t]{0.32\linewidth}
\centering
\includegraphics[width=0.95\textwidth]{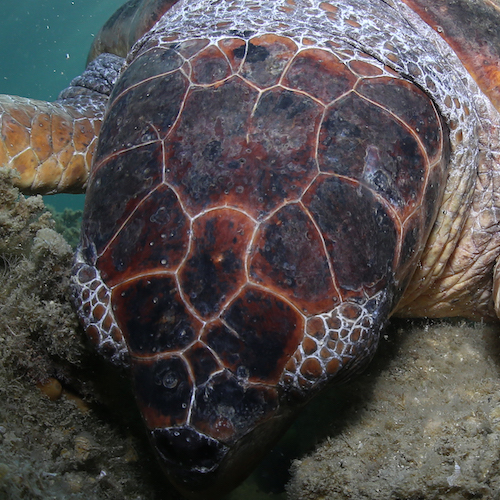}\\[0.5em]
\includegraphics[width=0.95\textwidth]{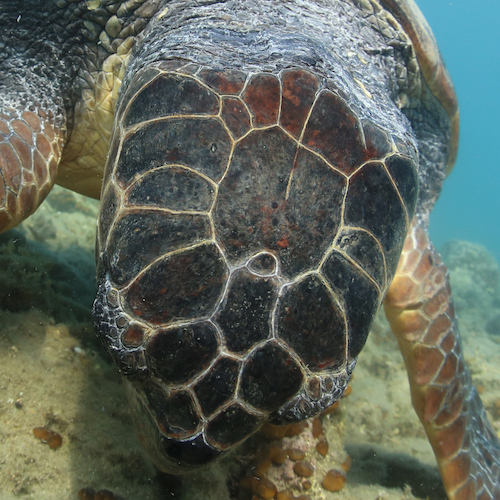}\\[0.5em]
\includegraphics[width=0.95\textwidth]{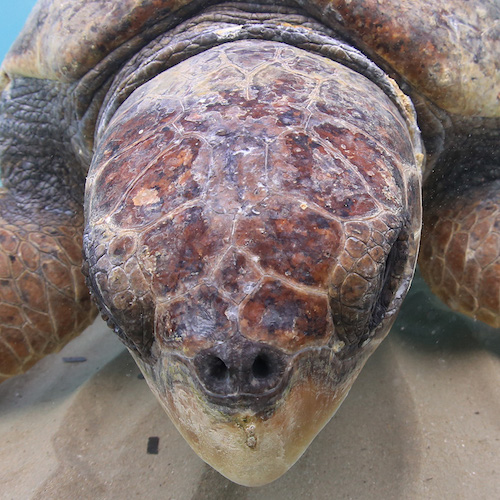}\\[0.5em]
\includegraphics[width=0.95\textwidth]{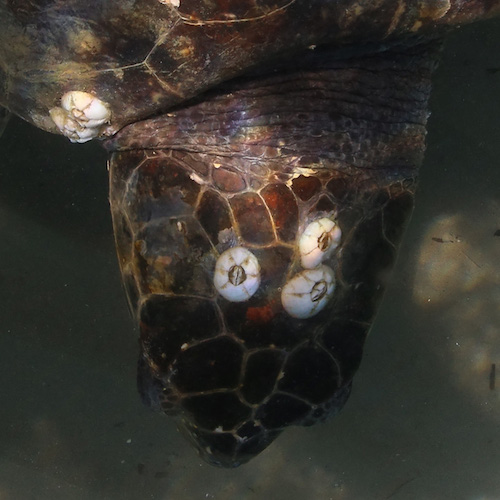}\\[0.5em]
\includegraphics[width=0.95\textwidth]{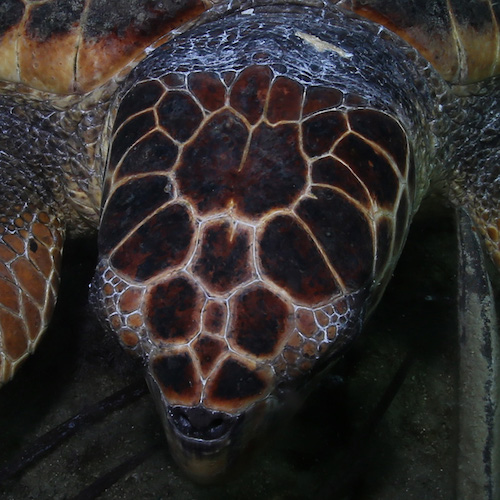}\\[0.5em]
\includegraphics[width=0.95\textwidth]{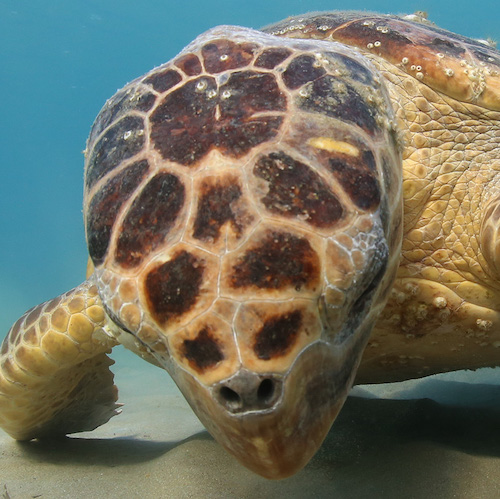}\\[0.5em]
\includegraphics[width=0.95\textwidth]{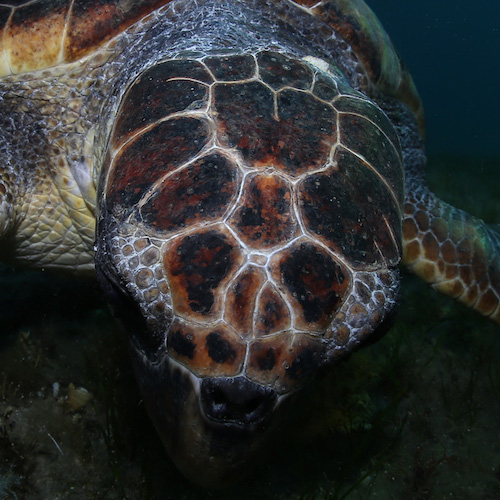}\\[0.5em]
\end{minipage}
\caption{Examples of 7 individuals (one individual per row) that show the variability of unique facial scale patterns of loggerhead sea turtles. From left to right: right lateral facial scales, left lateral facial scales, dorsal head scales.}
\label{fig:unique_patterns_app}
\end{figure}

\cref{fig:unique_patterns_app} displays photographs of seven individuals (one individual per row) showing the variability of the unique facial scale patterns of loggerhead sea turtles. The scales on the left and right sides of the head are different in a given individual, making it impossible to match them without any intermediate images.

\begin{figure}[t]
\centering
\begin{minipage}[t]{0.19\textwidth}
\centering
\includegraphics[width=0.95\textwidth]{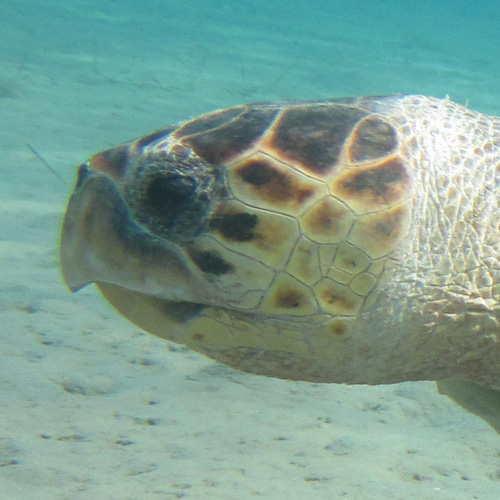}\\[-0.1em]
\small{\textbf{2011}}
\end{minipage}
\begin{minipage}[c]{0.05\textwidth}
\begin{center}
$\Large{\boldsymbol{\rightarrow}}$
\vspace{2.7cm}
\end{center}
\end{minipage}
\begin{minipage}[t]{0.19\textwidth}
\centering
\includegraphics[width=0.95\textwidth]{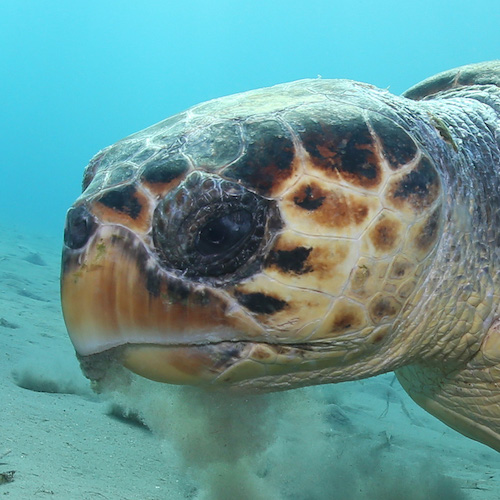}\\[-0.1em]
\small{\textbf{2019}}
\end{minipage}
\vspace{-1.8em}

\begin{minipage}[t]{0.19\textwidth}
\centering
\includegraphics[width=0.95\textwidth]{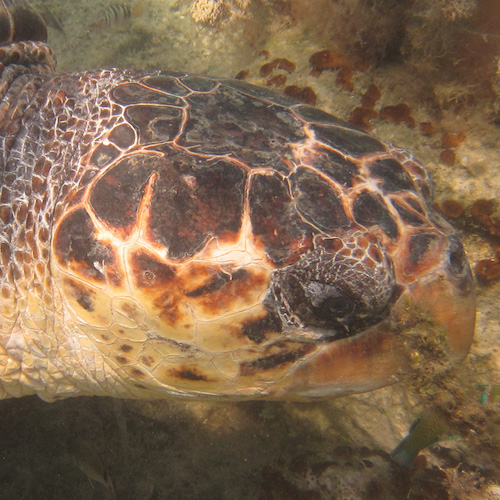}\\[-0.1em]
\small{\textbf{2011}}
\end{minipage}
\begin{minipage}[c]{0.05\textwidth}
\begin{center}
$\Large{\boldsymbol{\rightarrow}}$
\vspace{2.7cm}
\end{center}
\end{minipage}
\begin{minipage}[t]{0.19\textwidth}
\centering
\includegraphics[width=0.95\textwidth]{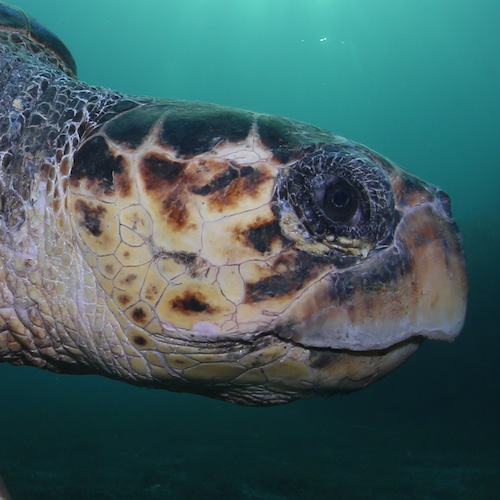}\\[-0.1em]
\small{\textbf{2019}}
\end{minipage}
\vspace{-1.8em}

\begin{minipage}[t]{0.19\textwidth}
\centering
\includegraphics[width=0.95\textwidth]{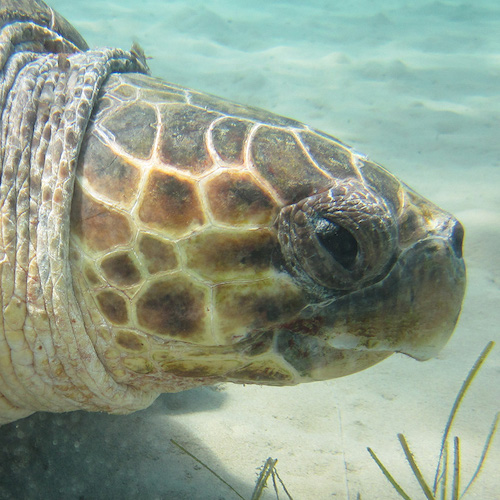}\\[-0.1em]
\small{\textbf{2012}}
\end{minipage}
\begin{minipage}[c]{0.05\textwidth}
\begin{center}
$\Large{\boldsymbol{\rightarrow}}$
\vspace{2.7cm}
\end{center}
\end{minipage}
\begin{minipage}[t]{0.19\textwidth}
\centering
\includegraphics[width=0.95\textwidth]{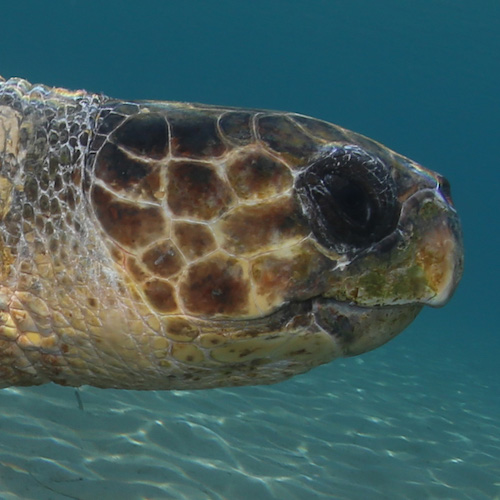}\\[-0.1em]
\small{\textbf{2021}}
\end{minipage}
\vspace{-1.8em}

\begin{minipage}[t]{0.19\textwidth}
\centering
\includegraphics[width=0.95\textwidth]{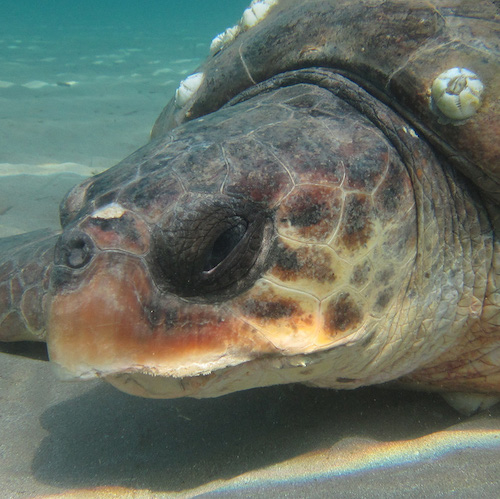}\\[-0.1em]
\small{\textbf{2013}}
\end{minipage}
\begin{minipage}[c]{0.05\textwidth}
\begin{center}
$\Large{\boldsymbol{\rightarrow}}$
\vspace{2.7cm}
\end{center}
\end{minipage}
\begin{minipage}[t]{0.19\textwidth}
\centering
\includegraphics[width=0.95\textwidth]{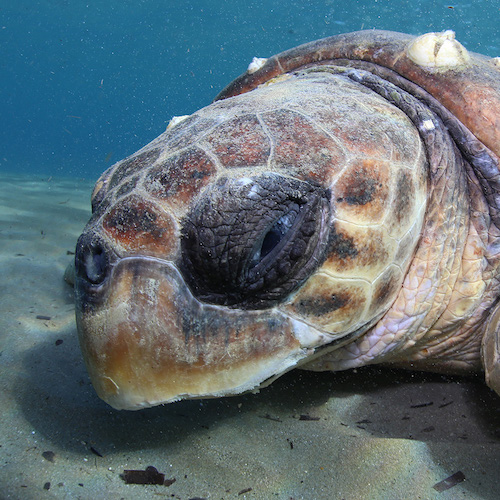}\\[-0.1em]
\small{\textbf{2019}}
\end{minipage}
\vspace{-1.8em}

\begin{minipage}[t]{0.19\textwidth}
\centering
\includegraphics[width=0.95\textwidth]{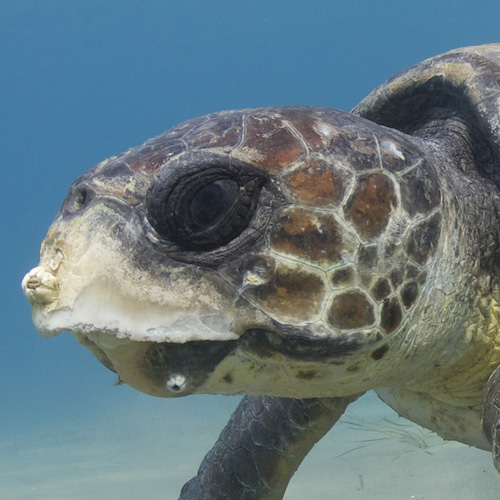}\\[-0.1em]
\small{\textbf{2014}}
\end{minipage}
\begin{minipage}[c]{0.05\textwidth}
\begin{center}
$\Large{\boldsymbol{\rightarrow}}$
\vspace{2.7cm}
\end{center}
\end{minipage}
\begin{minipage}[t]{0.19\textwidth}
\centering
\includegraphics[width=0.95\textwidth]{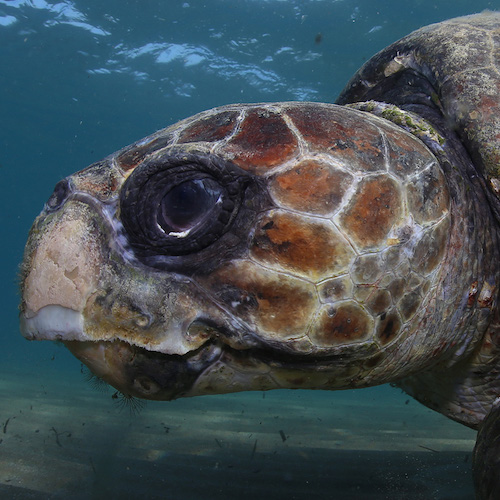}\\[-0.1em]
\small{\textbf{2019}}
\end{minipage}
\vspace{-2.5em}
\caption{Further examples of different visual appearances of the same individual sea turtles over long periods of time due to different factors like camera capture conditions and animal ageing. The shapes of the facial scales remained stable, but other features have changed over time, like colouration, pigmentation, shape, and scratches.}
\label{fig:changes}
\end{figure}

\cref{fig:changes} shows further examples of different visual appearances of the same individual sea turtles over long periods of time due to different factors like camera capture conditions and animal aging. The shapes of the facial scales remained stable, but other features have changed over time, like coloration, pigmentation, shape, and scratches.

\begin{figure*}[!ht]
\centering
\includegraphics[width=\linewidth]{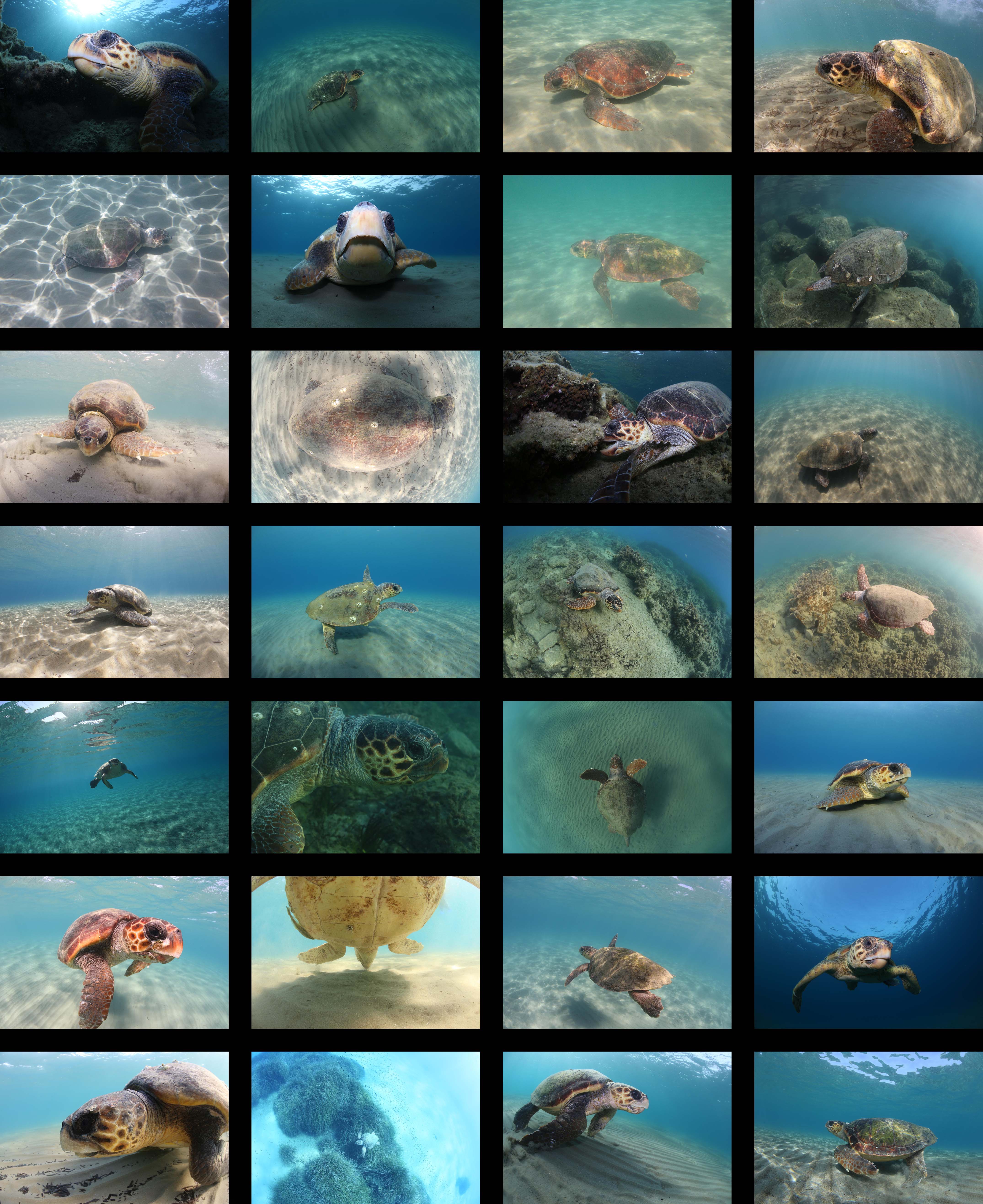}
\caption{Examples of original photographs from the SeaTurtleID2022 dataset.}
\label{fig:grid}
\end{figure*}

\cref{fig:grid} shows sample images from the SeaTurtleID2022 dataset, highlighting the variety of photographs (poses, orientations, backgrounds, etc.). 

\end{document}